\documentclass[10pt,twocolumn,letterpaper]{article}

\newcommand{\Tref}[1]{Table~\ref{#1}}
\newcommand{\Eref}[1]{Eq.~(\ref{#1})}
\newcommand{\Fref}[1]{Fig.~\ref{#1}}
\newcommand{\Sref}[1]{Sec.~\ref{#1}}

\newcommand*\Bell{\ensuremath{\boldsymbol\ell}}
\renewcommand{\paragraph}[1]{\vspace{1mm}\noindent\textbf{#1}}

\usepackage[pagenumbers]{cvpr} 

\usepackage{times}
\usepackage{epsfig}
\usepackage{graphicx}
\usepackage{amsmath}
\usepackage{amssymb}

\usepackage{booktabs}
\usepackage{array}
\usepackage{multirow}
\usepackage[table,xcdraw]{xcolor}
\usepackage{scalerel}

\definecolor{cvprblue}{rgb}{0.21,0.49,0.74}
\usepackage[pagebackref,breaklinks,colorlinks,citecolor=cvprblue]{hyperref}

\def\paperID{1170}
\def\confName{CVPR}
\def\confYear{2024}

\title{Attentive Illumination Decomposition Model \\ for Multi-Illuminant White Balancing}
\author{
Dongyoung Kim\textsuperscript{1} \qquad Jinwoo Kim\textsuperscript{1} \qquad Junsang Yu\textsuperscript{2} \qquad Seon Joo Kim\textsuperscript{1}\\
\textsuperscript{1}Yonsei University \qquad \textsuperscript{2}Samsung Advanced Institute of Technology\\
}

\begin{document}
\maketitle

\begin{abstract}
White balance (WB) algorithms in many commercial cameras assume single and uniform illumination, leading to undesirable results when multiple lighting sources with different chromaticities exist in the scene.
Prior research on multi-illuminant WB typically predicts illumination at the pixel level without fully grasping the scene's actual lighting conditions, including the number and color of light sources. This often results in unnatural outcomes lacking in overall consistency.
To handle this problem, we present a deep white balancing model that leverages the slot attention, where each slot is in charge of representing individual illuminants.
This design enables the model to generate chromaticities and weight maps for individual illuminants, which are then fused to compose the final illumination map.
Furthermore, we propose the centroid-matching loss, which regulates the activation of each slot based on the color range, thereby enhancing the model to separate illumination more effectively.
Our method achieves the state-of-the-art performance on both single- and multi-illuminant WB benchmarks, and also offers additional information such as the number of illuminants in the scene and their chromaticity. This capability allows for illumination editing, an application not feasible with prior methods.
\end{abstract}    
\section{Introduction}

\begin{figure}[t]
    \includegraphics[width=\linewidth]{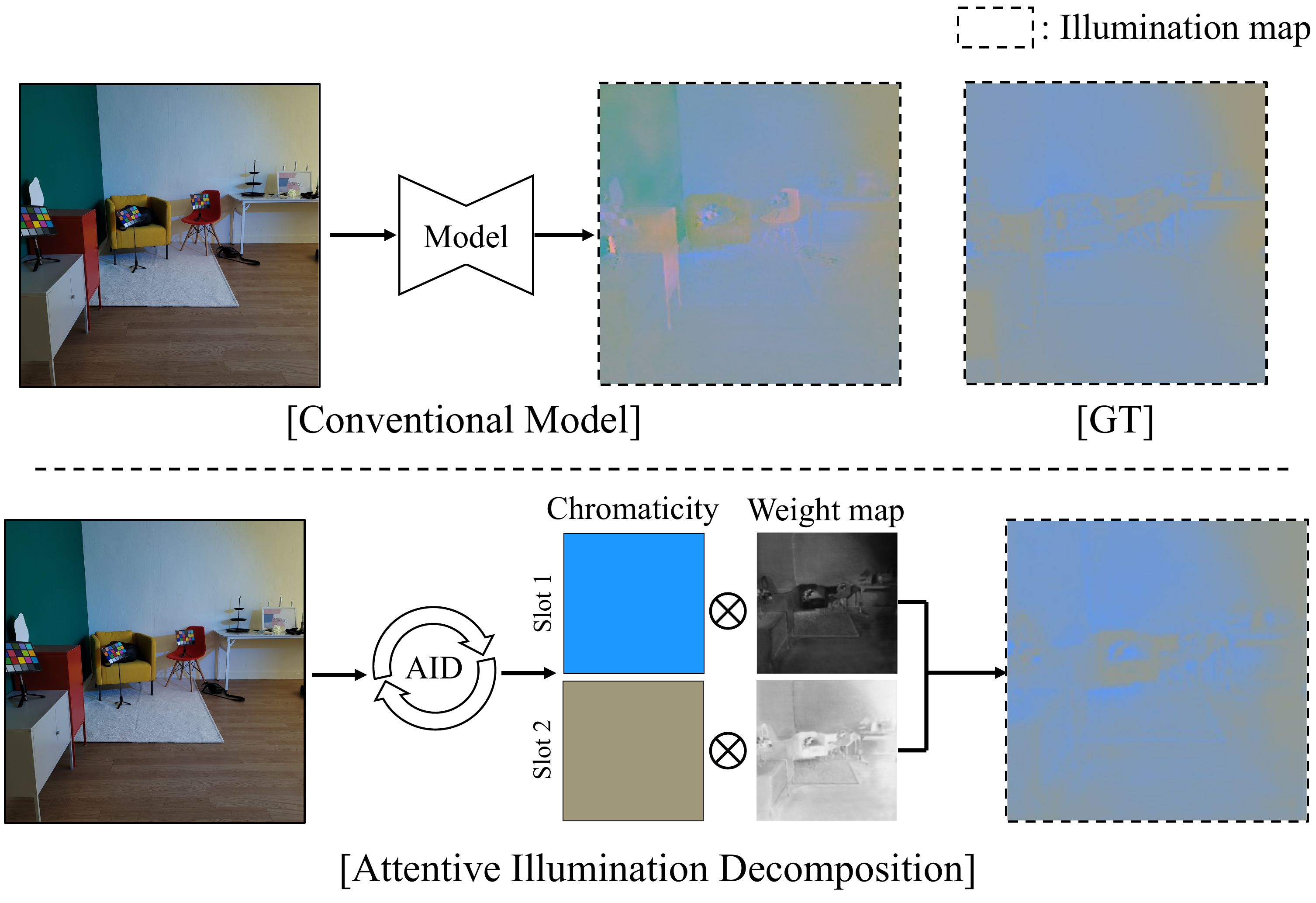}
    \vspace{-2em}
    \caption{Comparison of the AID framework (bottom) with existing approaches (top).
    Previous methodologies did not individually consider illuminant profiles within the scene, resulting in unnatural results.
   The AID framework outperforms previous works in illumination estimation  by estimating the chromaticity and pixel-wise weight map of each individual illuminant and combining them.}
\label{fig:teaser}
\end{figure}

Color constancy, a unique human capability, allows us to perceive the color of objects uniformly under any lighting conditions.
Similarly, a computational color constancy or white balancing (WB) module is integrated into the image processing unit, designed to compensate for the effects of illumination to recover the original color of the objects.

Many WB studies have been conducted with the goal of predicting the single chromaticity vector of the light source for a given image, assuming uniform illumination. Traditional statistics-based methodologies \cite{van2007edge, gijsenij2011improving, forsyth1990novel, gijsenij2010generalized}, including gray world \cite{buchsbaum1980spatial} and white patch \cite{land1977retinex} algorithms, used various statistics that could be obtained from images.
Data-driven methods \cite{barron2017fast, hu2017fc4} worked by optimizing through the white balance dataset.
However, these algorithms produce distorted results when multiple illuminants affect the scene simultaneously. For example, when a blue skylight is coming in from the window into a room with warm-colored lighting, applying a single white balance matrix to the entire image may fail in recovering the scene color.

Accordingly, spatially varying white balance algorithms have been proposed to deal with multi-illuminant scenes.
Early works estimate mixed illumination map by utilizing auxiliary flash photography \cite{hui2016white} or prior knowledge about the chromaticity of illuminants \cite{hsu2008light,beigpour2013multi}.
Recently, many DNN-based methods have been introduced with the advancements in neural networks.
Algorithms using patches \cite{bianco2017single}, GANs \cite{sidorov2019conditional}, U-Net \cite{kim2021large} with transformer blocks \cite{li2022transcc} have  been proposed.

All previous multi-illuminant WB works directly generate patch- or pixel-level predictions of illumination maps using an encoder-decoder structure without any structural constraints.
These approaches often fail to satisfy the linearity constraint \cite{hsu2008light, gijsenij2011color, kim2021large} that the chromaticity of mixed illumination can be expressed as a linear combination of individual light source chromaticity under the Lambertian image model.
This may result in producing unnatural illumination that does not exist in a scene (\Fref{fig:teaser} top).
In addition, as the previous methods cannot offer individual light source profiles in a multi-illuminant scene, further tuning or editing the illumination is not possible. 

To overcome the limitations of the existing multi-illuminant WB methods, we propose the Attentive Illumination Decomposition (AID) mechanism. AID shows strong performance and is equipped with tunability for the spatially varying multi-illuminant WB.
Our framework works in an end-to-end manner with a single given image. In other words, it does not require any auxiliary images \cite{hui2016white} or post-processing procedures \cite{hui2016white, hui2018illuminant} to decompose the illumination map.
Our model is based on slot attention~\cite{locatello2020object}, to learn the implicit representation of illuminant chromaticity in a scene in the form of slot vectors.
Specifically, we leverage the slot vectors to represent the chromaticities of the light sources in a scene, and use the attention map of each slot as the pixel-wise weight map of corresponding illuminant (\Fref{fig:teaser} bottom).
By doing so, we can enforce each predicted pixel-wise chromaticity to be a linear combination of the slot chromaticities, and enable illuminant-wise tunability.
The way our model generates the final illumination maps follows the linearity constraint so that our method can properly tackle the problem of spatially varying WB.
Furthermore, we propose a novel loss called centroid-matching loss, to effectively train our slot attention based model by assigning specific color ranges to slots.

We validate the robustness of AID framework through comprehensive experiments conducted on various datasets, including the LSMI dataset \cite{kim2021large}, the Multi-Illumination In the Wild dataset \cite{murmann2019dataset}, and the well-established single-illuminant dataset, NUS-8 \cite{cheng2014illuminant}. The experimental results consistently demonstrate superior performance compared to previous models, achieving the state-of-the-art performance across all of the aforementioned datasets.

Our contributions can be summarized as follows:
\begin{itemize}
    \item By successfully leveraging the concept of the slot attention, we propose a novel end-to-end framework AID, which can infer the chromaticities of illuminants and their pixel-level weight maps separately.
    \item
    We introduce the centroid-matching loss to enable more effective updates of slots to represent specific color gamuts.
    \item Our model not only demonstrates the state-of-the-art performance in both single- and multi-illuminant white balance scenarios but also provides tunable WB, thanks to its capacity to generate fully disentangled illumination maps.
\end{itemize}
\section{Related work}\label{2}
\subsection{Computational color constancy}\label{2.1}

\paragraph{Single-illuminant WB.}
Classical statistics-based algorithms utilizing image statistics have been studied \cite{buchsbaum1980spatial, land1977retinex, finlayson2004shades, van2007edge} for computational color constancy. Additionally, numerous WB datasets \cite{ciurea2003large, gehler2008bayesian, shi2000re, cheng2014illuminant} have been proposed for data-driven research. Methodologies have been introduced involving the learning of kernels to detect illuminant chromaticity in the uv-histogram space \cite{barron2015convolutional, barron2017fast}, utilizing convolutional features \cite{bianco2015color, shi2016deep, hu2017fc4, oh2017approaching}, and employing various learning techniques \cite{lo2021clcc, xu2020end, qian2017recurrent, yu2020cascading}.
In particular, FC4 \cite{hu2017fc4} employs a form of attention technique by inferring spatial weighting coefficients, rather than uniformly using all spatial features within the image. On the other hand, C4 \cite{yu2020cascading} demonstrated the capability for more accurate chromaticity inference through iterative refinement process. While they achieve impressive results for single-illuminant WB, they cannot address the multi-illuminant WB cases.
We found that the incorporation of spatial attention maps and an iterative refinement strategy, in conjunction with the concept of slots outlined in \Sref{2.2}, is highly suitable for addressing the spatially varying multi-illuminant decomposition task.

\paragraph{Multi-illuminant WB.}
To solve the multi-illuminant WB problem, several studies have been conducted to utilize additional prior information such as the chromaticity of the illuminant \cite{hsu2008light, beigpour2013multi}, flash photography \cite{hui2016white}, and human face \cite{bianco2014adaptive}. 
Approaches that apply single-illuminant WB in a spatially varying form have been introduced in~\cite{bleier2011color, gijsenij2011color}, and a graph structure reflecting the characteristics of spatially varying WB has been utilized in \cite{mutimbu2016multiple}.

Following small-scale multi-illuminant datasets \cite{bleier2011color, gijsenij2011color, beigpour2013multi, bianco2017single} for testing spatially varying WB algorithms, several large scale multi-illuminant datasets have been captured \cite{murmann2019dataset, kim2021large} and synthesized \cite{hao2020multi} recently.
Deep learning-based strategies such as using GANs \cite{sidorov2019conditional}, and leveraging transformer blocks with multi-task learning strategies \cite{li2022transcc} have also been explored. 

The base architecture for previous multi-illuminant WB used the encoder-decoder structure to directly predict the chromaticity of illumination for each individual pixel.
These models fall short in estimating and incorporating the individual chromaticities of illuminants present in the scene, leading to inconsistencies in the generated illumination map. (Fig.~\ref{fig:teaser} top).
While a model that estimates pixel-wise weights for pre-specified WB presets \cite{afifi2022auto} has been proposed recently, the resulting weight maps do not accurately reflect the the ground-truth illuminant-wise mixing ratio due to its reliance on pre-defined WB presets.
A summary of the comparison between our framework and previous works is presented in Table~\ref{tab:model_ability}.

\begin{table}
\begin{center}
\resizebox{\linewidth}{!}{
\renewcommand{\arraystretch}{1.1}
\begin{tabular}{cccc}
\toprule
Models & \begin{tabular}[c]{@{}c@{}} Mixed \\ illumination \end{tabular} &  \begin{tabular}[c]{@{}c@{}} Decomposed \\ illumination map \end{tabular}  & \begin{tabular}[c]{@{}c@{}} Controllable \\ WB \end{tabular} \\ \midrule \midrule
Single AWB & $\times$ &  $\times$ & $\times$  \\ \midrule
\begin{tabular}[c]{@{}c@{}} Multi-AWB \\ \cite{afifi2022auto,sidorov2019conditional,li2022transcc} \end{tabular} & $\checkmark$  & $\times$ & $\times$ \\ \midrule
\textbf{AID (Ours)} &\textbf{$\checkmark$} & \textbf{$\checkmark$} &  \textbf{$\checkmark$} \\ \bottomrule
\end{tabular}}
\end{center}
\vspace{-1.5em}
\caption{Comparison between previous WB methods and our AID framework. 
AID predicts a decomposed illumination map, enabling the inference of individual illuminant chromaticity and the number of illuminants in a scene. This new feature enables controllable WB, allowing for individual adjustment of the color of each illuminant in a scene.}
\vspace{-1.5em}
\label{tab:model_ability}
\end{table}

\subsection{Slot attention}\label{2.2}
Slot attention \cite{locatello2020object} was introduced to solve the object-centric learning (OCL), where the model needs to cluster and compute the representation of objects from a given scene without any human-annotated labels in an autoencoding manner. 
Slot attention employs the concept of the slots, 
a set of vectors, each of which captures the representation of the object in a scene.
Slots are initialized using Gaussian random sampling and are subsequently evolved to capture the representations of objects.
Dot-product based attention maps between slots and encoded visual feature maps are used for updating the slots.
By applying slot-wise softmax mechanism on the attention map, slots are forced to compete with each other to get more task-relevant representation, i.e. object-centric representation.

Due to the decomposition ability of the slot attention, it has been widely applied to various domains in computer vision such as object discovery \cite{locatello2020object, engelcke2019genesis, kipf2021conditional, kim2023shepherding}, novel view synthesis \cite{sajjadi2022object}, panoptic segmentation \cite{zhou2022slot}, and visual question answering \cite{wang2020language}.
Slot attention acts like a soft k-means clustering, where the slots are  appropriately updated to represent the target sub-element.
In this work, we adopt slot attention to the task of multi-illuminant white balancing, enforcing slots to implicitly represent individual illuminants.
In addition, we introduce a novel loss function named centroid matching loss, aimed at preventing all slots from indiscriminately contributing to the inference.
This improves illumination decomposition accuracy by allocating the specific color ranges to each slot.
\section{Method}

\begin{figure*}[t]
    \includegraphics[width=\textwidth]{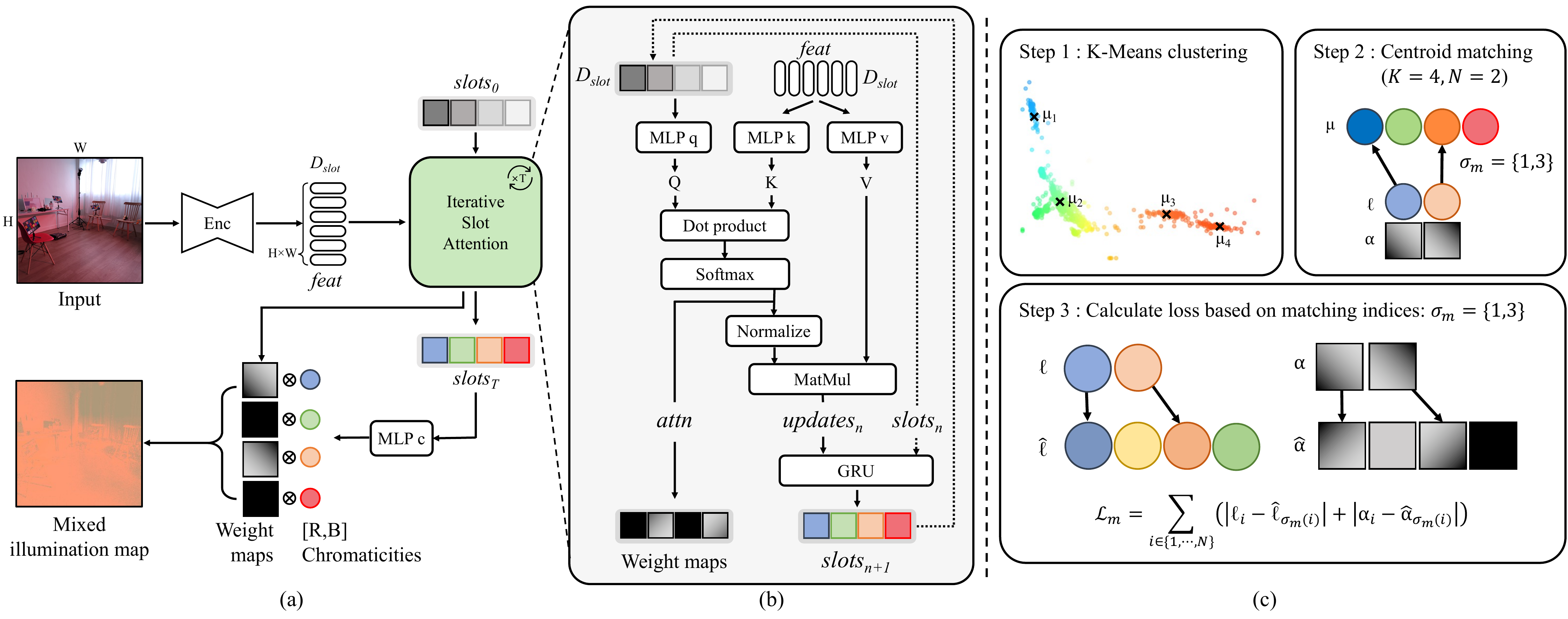}
    \vspace{-2em}
    \caption{(a) Overview of our framework. Image feature is extracted from the input using an U-Net encoder. Next, the slot attention adaptively calibrates slot representation to be bound with illuminant chromaticity in each scene. Finally, the model fuses the chromaticity and the weight map to generate the mixed illumination map. (b) Detailed generation flow of weight maps and calibrated slots, where Q-Softmax denotes softmax application on the query dimension. (c) Illustration of the slot-wise loss using the centroid based Hungarian matching under $K=4, N=2$ assumption.}
\label{fig:model_overview}
\end{figure*}

\subsection{Image formation model}
In the Lambertian image model, the RGB value of each pixel under single-illuminant condition can be represented as follows:
\begin{equation}
    I = kR \circ \Bell.
\end{equation}
Here $I$, $R$ and $\Bell$ are $3 \times 1$ vectors for observed RGB pixel, surface reflectance, and normalized illuminant chromaticity, respectively. The $\circ$ symbol represents element-wise product and the scalar $k$ represents the integrated scaling term of illumination including the power of illuminant and surface normal.
In this paper, we normalize the illuminant chromaticity so that the value of the green channel becomes 1.
Previous works \cite{hsu2008light, gijsenij2011color, kim2021large} suggest that if multiple illuminants are present in a scene, the chromaticity of mixed illumination can be represented by the linear combination of the chromaticity of each illuminant. This property has been used to calculate per-pixel illumination labels for multi-illuminant datasets \cite{beigpour2013multi, kim2021large}.

Under the imaging model, the illumination chromaticity value $\Bell$ on a given location $x$ of a single or multiple illuminant scene can be generalized and expressed as follows:
\begin{equation}
\begin{aligned}
    \Bell_{mixed}(x) = &\sum_{k=1}^{N} \alpha_k(x) \Bell_k,  \\
    \textrm{where} \quad  &\sum_{k=1}^{N} \alpha_k(x) = 1, \quad 
    \Bell_k = \begin{bmatrix}
    R_k \\
    B_k \end{bmatrix} .
\label{eq1:image_model}
\end{aligned}
\end{equation}
$\alpha_k$ and $\Bell_k$ represent the weight map and the normalized chromaticity of illuminant $k$, respectively, and $N$ is the number of illuminants in the scene.
As mentioned earlier, we only consider the R and B channels of the illuminant chromaticity $\Bell_k$, given that the G channel is normalized to 1.
Since $\Bell_k$ is the chromaticity of each light source, it does not change with pixel location, and only the weight map $\alpha$ is dependent on $x$.

\subsection{Attentive illumination decomposition}
To solve the multi-illuminant WB, we design the Attentive Illumination Decomposition (AID) framework, which follows the imaging model described in \Eref{eq1:image_model}.
The proposed method first predicts the weight map $\alpha_k$ and the chromaticity $\Bell_k$ of each illuminant in the scene, and then the mixed illumination map $\Bell_{mixed}$ for WB is generated using \Eref{eq1:image_model}.
It is important to highlight that this is the first approach to separately predict the chromaticity and the weight map of illuminant for multi-illuminant WB, leading to enhancement in performance.

To obtain $\alpha_k$ and $\Bell_k$, our framework utilizes the slot attention \cite{locatello2020object}.
Different from the existing slot attention models, where slots are typically designed to capture object-level features, we design the model to make the slots to represent illuminant-level information.
More precisely, each slot in our model allows us to infer both the chromaticity and the weight map associated with the corresponding illuminant.
The overview of our framework is illustrated in \Fref{fig:model_overview}(a).
Our model consists of three parts: 1) image feature extraction, 2) iterative slot calibration process using slot attention, and 3) weight map \& illuminant chromaticity fusion.

\paragraph{Image feature extraction.}
For a given raw image $\mathbf{I}$ with the resolution $H \times W$, the feature encoder $E$ extracts a latent feature $feat$ with the same spatial resolution as the image and $D_{slot}$ channels:
\begin{equation}
    feat := E(\mathbf{I}) \in \mathbb{R}^{HW \times D_{slot}},
\label{eq2:feature_extraction}
\end{equation}
where $D_{slot}$ represents the dimension of slots.

\paragraph{Iterative slot calibration by slot attention.}
After extracting the image features, the iterative slot attention module (\Fref{fig:model_overview}(b)) is applied to calibrate representations of the illuminant chromaticity.
The details of the calibration process are as follows.

First, $slots_{\scaleto{0}{3.1pt}}$ $\in \mathbb{R}^{K \times D_{slot}}$ are initialized as learnable parameters where $K$ indicates the number of slots. 
The slot attention module takes the image feature $feat$ and the initialized $slots$ as inputs to produce attention map $attn$:
\begin{equation}
\begin{gathered}
    attn_{i,j} := \frac{\mathrm{exp}(M_{i,j})}{\sum_l{\mathrm{exp}(M_{i,l})}}, \quad \textrm{where} \\
    M := \frac{1}{\sqrt{D_{slot}}}k(feat) \cdot q(slots_{\scaleto{n}{3pt}})^T \in \mathbb{R}^{HW \times K},
\label{eq3:attention_softmax}
\end{gathered}
\end{equation}
where $k$ and $q$ are MLPs for generating the key and query representations in $D_{slot}$ dimension, and $slots_{n}$ represents the state of the slots in the n-th iteration.

Subsequently, the intermediate representation vectors, $updates$, are computed by aggregating the $values$ through spatially normalized attention map $W$:
\begin{equation}
\begin{gathered}
    updates := W^T \cdot v(feat) \in \mathbb{R}^{K \times D_{slot}}, \\
    \textrm{where} \quad W_{i,j} := \frac{attn_{i,j}}{\sum_{l=1}^{N}attn_{l,j}},
\label{eq4:update_calculation}
\end{gathered}
\end{equation}
where $v$ is MLPs for generating value representation. 

Finally, the calibrated $slots_{n+1}$ are refined by the GRU \cite{chung2014empirical}, which takes $updates$ as input and previous $slots_{n}$ as hidden state:
\begin{equation}
\begin{gathered}
slots_{\scaleto{n+1}{5pt}} = GRU(slots_{{\scaleto{n}{3pt}}},updates_{\scaleto{n}{3pt}}).
\end{gathered}
\label{eq5:gru}
\end{equation}
The process from \Eref{eq3:attention_softmax} to \Eref{eq5:gru} is repeated $T$ times to generate final calibrated slots, $slots_{T}$.

\begin{figure}[t]
    \includegraphics[width=\linewidth]{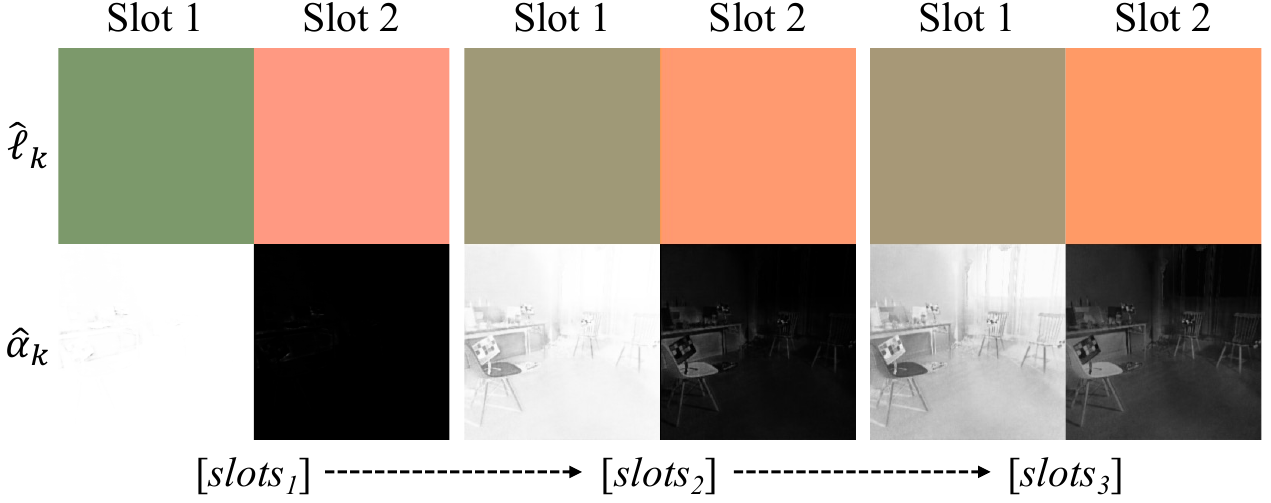}
    \caption{Slot calibration process. The chromaticity $\ensuremath{\boldsymbol{\hat{\ell}}}_k$ and weight map $\hat{\alpha}_k$ generated from each $slots_n$ are iteratively calibrated to their ground truth values.}
    \vspace{-1em}
\label{fig:slot_evolution}
\end{figure}

\paragraph{Weight map \& Illuminant chromaticity fusion.}
In our framework, we have carefully designed the output tensors of the slot-attention module, $slots$ and $attn$, to represent the chromaticity of illuminants $\Bell$ and the weight map $\alpha$, respectively.
Specifically, the $HW \times K$ shaped tensor $attn$, represents the pixel-wise similarity score between $feat$ and each $slot$, enabling its direct use as the set of $K$ weight maps for each illuminant ($\hat{\alpha}_{1} \dots \hat{\alpha}_{K}$).
Also, chromaticities for $K$ illuminants ($\ensuremath{\boldsymbol{\hat{\ell}}}_{1} \dots \ensuremath{\boldsymbol{\hat{\ell}}}_{K}$) can be generated through passing calibrated $slots_{\scaleto{T}{4pt}}$ to chromaticity conversion MLPs $c$, where $c(slots_{\scaleto{T}{4pt}})$ is a $K \times 2$ shaped tensor.
Finally, we can fuse these two tensors to make mixed illumination map $\ensuremath{\boldsymbol{\hat{\ell}}}_{mixed}$ according to \Eref{eq1:image_model}, by simply multiplicating them:
\begin{equation}
\begin{gathered}
    \ensuremath{\boldsymbol{\hat{\ell}}}_{mixed} = \sum_{k=1}^{K} \hat{\alpha}_k \ensuremath{\boldsymbol{\hat{\ell}}}_k = attn \cdot c(slots_T).
\label{eq6:fusion}
\end{gathered}
\end{equation}
Although the above equation utilizes the final calibrated slots, $slots_T$, it is noteworthy that we can visualize the chromaticity $\ensuremath{\boldsymbol{\hat{\ell}}}_{k}$ and weight map $\hat{\alpha}_k$ for each iteration by employing the respective $slots_n$ of iteration $n$.
\Fref{fig:slot_evolution} demonstrates how the generated illuminant chromaticity $\ensuremath{\boldsymbol{\hat{\ell}}}_{k}$ and weight map $\hat{\alpha}_k$ changes as $slots$ are itaratively calibrated.

\subsection{Loss functions}
AID framework is trained with two types of loss function: 1) mixed illumination loss, and 2) slot-wise loss, named as centroid-matching loss. The total training objective is to minimize the sum of these two loss functions.

\paragraph{Mixed illumination loss.}
Mixed illumination loss $\mathcal{L}_{mixed}$ is simply defined by L1 distance between the predicted mixed illumination map $\ensuremath{\boldsymbol{\hat{\ell}}}_{mixed}$ and the ground truth $\Bell_{mixed}$:
\begin{equation}
    \mathcal{L}_{mixed} = \left| \Bell_{mixed} - \ensuremath{\boldsymbol{\hat{\ell}}}_{mixed} \right| .
\end{equation}

\paragraph{Centroid-based matching loss.}
The mixed illumination loss alone does not provide a sufficient constraint that ensures our model activates the appropriate number of slots. 
Instead, it may result in the activation of either all slots or a random number of slots.
As depicted in \Fref{fig:model_overview}(a), the scene involves two illuminants, yet the model employs four slots to generate a mixed illumination map. Hence, it is necessary to strategically select and supervise the slots which are aligned with the ground-truth chromaticity and weight map.
In this context, we design the loss term based on two assumptions that 1) each slot possesses its pre-defined cluster (color-gamut), and 2) activation of the slot should occur when the ground truth chromaticity lies within its cluster boundary.
To this end, we propose the centroid matching loss and the calculation process of this loss is presented in \Fref{fig:model_overview}(c).

First let us denote a pre-calculated set of centroids as $\mu = \{\mu_i\}_1^K$, obtained by applying K-means algorithm on the illuminant chromaticity distribution of the dataset. These centroids serve as the centerpoints of each illuminant chromaticity cluster, and in AID framework, each slot is responsible for representing one of these clusters. Next, we obtain a set of matched indices $\sigma_m$ that minimizes the L1 cost between the matched centroid chromaticitiy $\mu$ and the ground truth chromaticity $\Bell$:
\begin{equation}
    \sigma_m = \underset{\sigma}{argmin} \sum_i^N \left| \Bell_{i} - \mu_{\sigma(i)} \right|,
\end{equation}
where N is the number of ground-truth illuminants in each scene and $\sigma$ is one of the combinations of N elements from the set $\{ 1, \cdots ,K \}$.
Now we can define the loss term with respect to the chromaticity and weight map of each matched slots as follows:
\begin{equation}
\begin{gathered}
    \resizebox{.88\hsize}{!}{
    $\mathcal{L}_{centroid} = \sum_{i} \left(\left| \Bell_i - \ensuremath{\boldsymbol{\hat{\ell}}}_{\sigma_m(i)} \right|
    + \left| \alpha_i - \hat{\alpha}_{\sigma_m(i)} \right| \right)$,}
\end{gathered}
\end{equation}
where the centroid-matching loss $\mathcal{L}_{centroid}$ consists of both L1 loss for chromaticity and weight map of the matched slot indices.
Here, the predicted weight map $\hat{\alpha}_k$ and the illuminant chromaticity $\hat{\Bell}_k$, are the k-th channel of $attn$ and $c(slots_T)$, as previously shown in \Eref{eq6:fusion}.
\section{Experiments}
\subsection{Experimental setup}
We validate the multi-illuminant WB performance of AID using two datasets: the LSMI dataset \cite{kim2021large} captured with three cameras having different bit-depths and spectral sensitivities, and the Multi-Illumination In the Wild dataset (MIIW) \cite{murmann2019dataset}, which is a versatile dataset covering various illumination-related tasks, including multi-illuminant WB.

We use seven slots ($K=7$), 64 latent channels for $D_{slot}$, and calibrate slots three times ($T=3$).
For the evaluation, the green channel was inserted (G=1) to the mixed illumination map $\ensuremath{\boldsymbol{\hat{\ell}}}_{mixed}$, and the mean angular error (MAE) in degree was calculated with respect to the ground truth illumination map.
For more detailed information, please refer to the supplementary materials.

\subsection{Spatially varying white balance}

\begin{table}
\begin{center}
\resizebox{\linewidth}{!}{
\begin{tabular}{llcccccc}
\toprule
&              & \multicolumn{3}{c}{mean}                      & \multicolumn{3}{c}{median}        \\ \midrule \midrule
\multirow{7}{*}{\begin{tabular}[c]{@{}l@{}}(a)\end{tabular}} & AWB \cite{afifi2019color} \textdagger           & \multicolumn{3}{c}{9.54}                      & \multicolumn{3}{c}{8.19}          \\
                                                                                          & Patch CNN  \cite{bianco2017single}    \textdagger & \multicolumn{3}{c}{4.82}                      & \multicolumn{3}{c}{4.24}          \\
                                                                                          & AngularGan \cite{sidorov2019conditional} \textdagger  & \multicolumn{3}{c}{4.69}                      & \multicolumn{3}{c}{3.88}          \\
                                                                                          & TransCC \cite{li2022transcc} \textdagger     & \multicolumn{3}{c}{2.78}                      & \multicolumn{3}{c}{2.15}          \\
                                                                                          & LSMI-U \cite{kim2021large}        & \multicolumn{3}{c}{2.31}                      & \multicolumn{3}{c}{1.89}          \\
                                                                                          & AID          & \multicolumn{3}{c}{\textbf{2.04}}             & \multicolumn{3}{c}{\textbf{1.73}} \\ 
                                                                                          & AID + MDL & \multicolumn{3}{c}{\textbf{1.93}}             & \multicolumn{3}{c}{\textbf{1.60}} \\ \bottomrule
                                                                                          &              & \multicolumn{3}{c}{mean}                      & \multicolumn{3}{c}{median}        \\ \cmidrule(lr){3-5} \cmidrule(lr){6-8}
                                                                                          &              & gal.          & son.          & nik.          & gal.       & son.      & nik.     \\ \midrule \midrule
\multirow{3}{*}{(b)}                                                          & LSMI-H \cite{kim2021large}      & 3.06          & 3.21          & 2.99          & 2.54       & 2.89      & 2.61     \\
                                                                                          & LSMI-U \cite{kim2021large}       & 2.68          & 2.15          & 1.92 & 2.17       & 1.74      & 1.54     \\
                                                                                          & AID          & \textbf{1.66} & \textbf{1.66} & \textbf{1.71}          & \textbf{1.41}       & \textbf{1.35}      & \textbf{1.34}     \\ \bottomrule
                                                                                          
\end{tabular}
}
\end{center}
\vspace{-1.5em}
\caption{Mean angular error (MAE) for the spatially varying illumination map on LSMI dataset: (a) all-in-one (cross-camera), (b) device-specific.
{\textdagger} indicates that the results of \cite{li2022transcc} are referenced.
}
\label{tab:multi-wb}
\end{table}

\begin{table}
\begin{center}
\resizebox{\linewidth}{!}{
\begin{tabular}{ccccccc}
\toprule
\multirow{2}{*}{\textbf{LSMI \cite{kim2021large}}} & \multicolumn{2}{c}{Galaxy} & \multicolumn{2}{c}{Sony} & \multicolumn{2}{c}{Nikon} \\
              \cmidrule(lr){2-3} \cmidrule(lr){4-5} \cmidrule(lr){6-7}
              & single       & multi       & single      & multi      & single      & multi       \\ \midrule \midrule
LSMI-U \cite{kim2021large}    & 2.85      & 2.55     & 1.92     & 2.34    & 1.49     & 2.30     \\
AID & \textbf{1.19} & \textbf{2.03} & \textbf{1.01} & \textbf{2.16} & \textbf{1.11} & \textbf{2.26} \\ \bottomrule
\end{tabular}}
\end{center}
\vspace{-1.5em}
\caption{Average MAE values obtained through experiments on the LSMI test set, distinguishing between single and multi-illuminant scenarios.}
\label{tab:single_multi}
\end{table}

\begin{table}
\begin{center}
\resizebox{\linewidth}{!}{
\begin{tabular}{ccccccc}
\toprule 
\multirow{2}{*}{\textbf{MIIW \cite{murmann2019dataset}}} &  \multicolumn{2}{c}{Single (1)} & \multicolumn{2}{c}{Multi (2,3)} & \multicolumn{2}{c}{Mixed (1,2,3)} \\ 
\cmidrule(lr){2-3} \cmidrule(lr){4-5} \cmidrule(lr){6-7}
& mean & median & mean & median & mean & median \\ \midrule \midrule
LSMI-U \cite{kim2021large}  & 4.15& 2.39& 4.34& 3.87& 4.28& 3.54\\ 
AID & \textbf{1.07} & \textbf{0.73} & \textbf{3.14} & \textbf{2.81} & \textbf{2.46} & \textbf{2.11} \\
\bottomrule
\end{tabular}
}
\end{center}
\vspace{-1.5em}
\caption{MAE values for predicting single-, multi-, and mixed-illuminant scenario using the MIIW test set.}
\label{tab:miiw}
\end{table}

\paragraph{Quantitative comparison.}
For the LSMI dataset, we evaluated AID under two settings to show the robustness of the proposed method: all-in-one cross-camera and device-specific setting.
As shown in \Tref{tab:multi-wb}(a), AID achieves the state-of-the-art performance compared to all previous models in LSMI dataset.
Here, we would like to inform that LSMI-H and LSMI-U are the preceding state-of-the-art models introduced in the LSMI dataset. 
LSMI-H employs HDR-Net \cite{gharbi2017deep}, while LSMI-U utilizes U-Net \cite{ronneberger2015u}.

Moreover, as AID uses a concept of slots as an intermediate representation of the illumination, the model can be easily extended to multi-domain learning (MDL).
We simply make the slot initialization different depending on the camera model (AID + MDL) and this slight modification brings additional 5\% performance enhancement.
Furthermore, in the camera-specific setting (\Tref{tab:multi-wb} (b)), AID outperforms the LSMI baselines for all three cameras.

Since the LSMI dataset consists of  one- to three-illuminant scenes, we also tested device-specific models with single and multi (two to three) illuminant subset, separately. As illustrated in \Tref{tab:single_multi}, our framework consistently outperforms the LSMI baseline \cite{kim2021large} across all devices in both single- and multi-illuminant settings.
For the single-illuminant case, we could observe that only one slot is activated among 7 slots, and produces near-perfect global uniform illumination, resulting in a significant performance improvement over LSMI baseline.

We further demonstrate the robustness of our framework using another large-scale dataset, Multi-Illumination in the Wild \cite{murmann2019dataset}.
Since no other algorithms have been applied to MIIW dataset previously, we select LSMI-U as our baseline.
\Tref{tab:miiw} demonstrates that AID outperforms LSMI-U, which had previously shown the best performance on the LSMI dataset, further highlighting the superior performance of AID.

\begin{figure*}[t]
    \includegraphics[width=\textwidth]{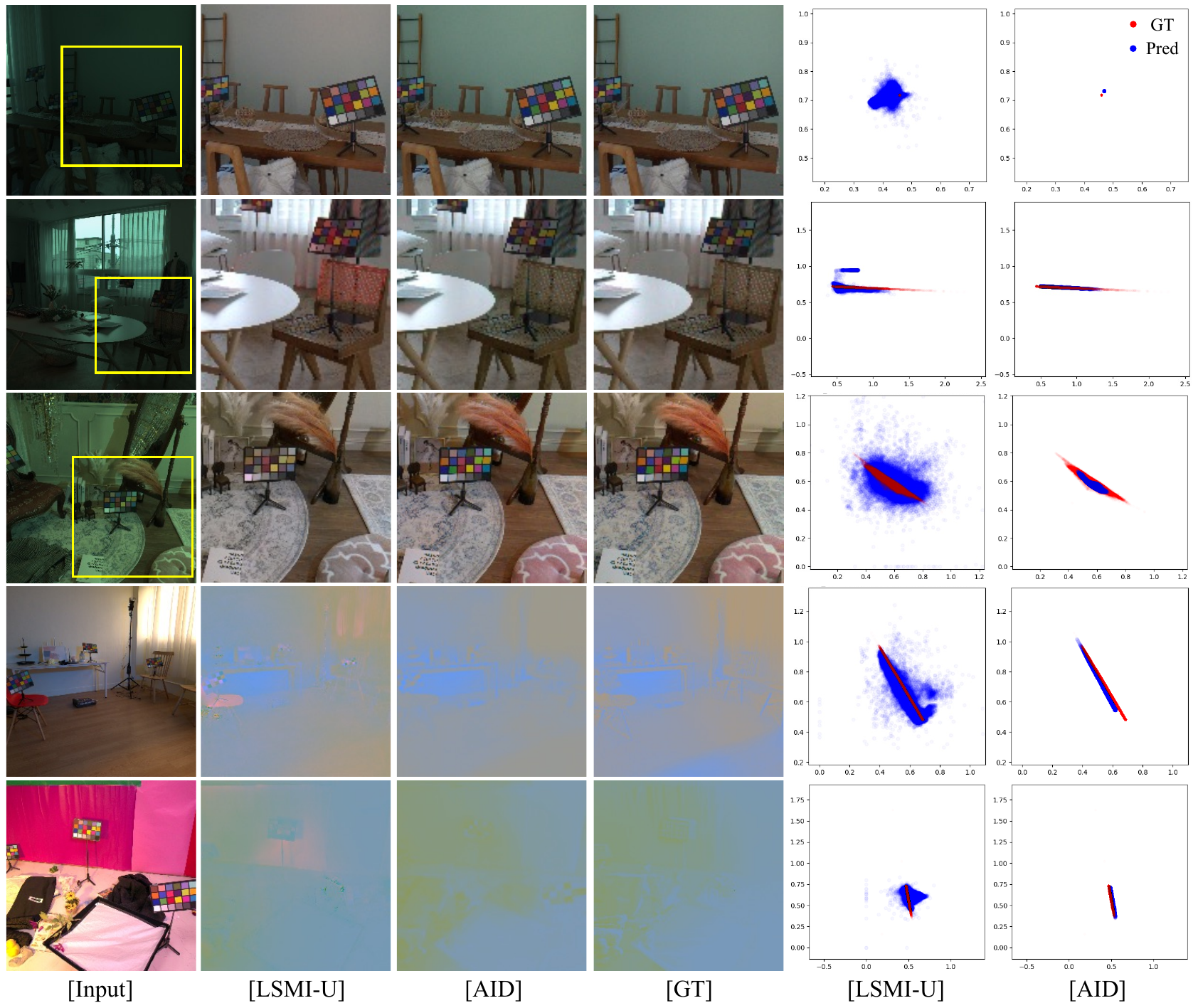}
    \caption{Qualitative comparison using LSMI test set. Top three rows show original raw image and corresponding WB results. The last two rows show the sRGB input images and corresponding illumination maps. The two rightmost columns demonstrate that our model, which infers illuminant-wise chromaticity and spatially mixes them, leads to more stable illumination plots compared to previous approaches. The x-axis and y-axis of the plot represent the ratio of the illumination value of the R and B channels to the value of the G channel.}
\label{fig:result_comparison}
\end{figure*}

\begin{figure*}[t]
    \includegraphics[width=\textwidth]{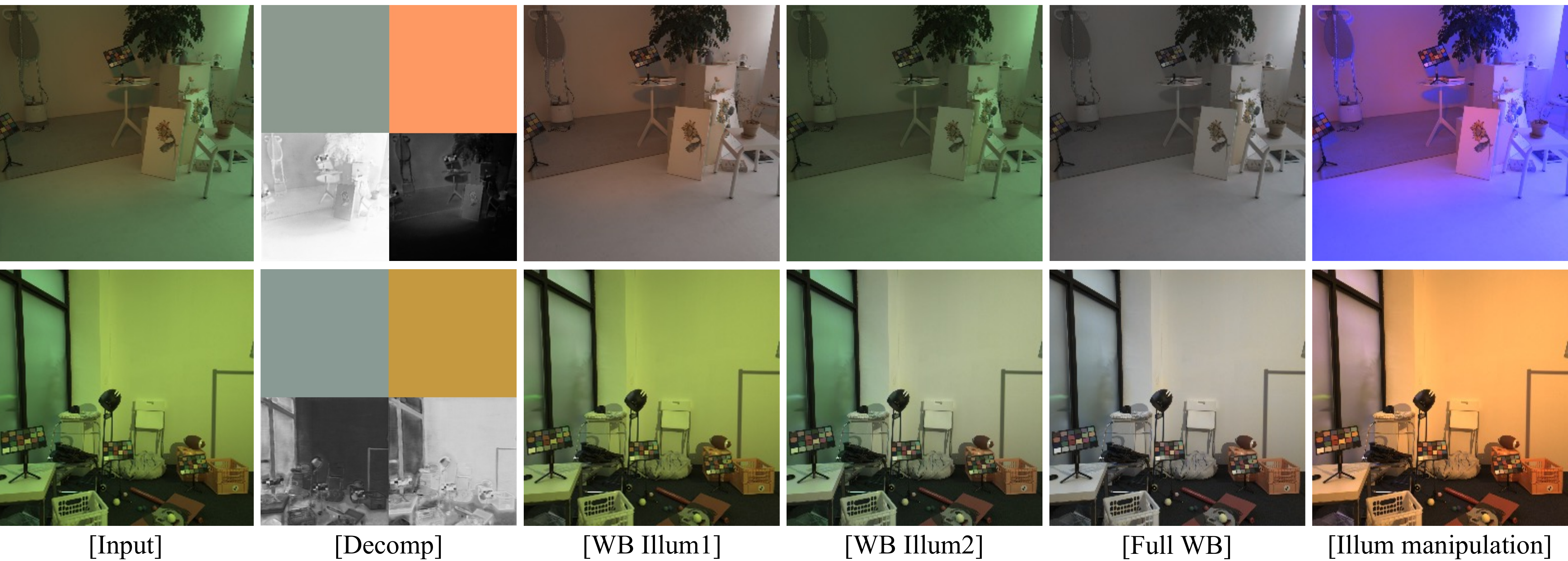}
    \caption{Further applications of AID framework on LSMI test set examples. The separated weight map and the corresponding illuminant chromaticity (Decomp) allow for individual white balance to be applied to each light (WB Illum1,2), and for the chromaticity to be adjusted as desired (Illum manip). Full WB shows the results of applying white balance to all illuminants for reference. Gamma was adjusted for all images to increase visibility, and the G channel was scaled down for the decomposed illumination map visualization.}
\label{fig:selective_awb}
\end{figure*}

\paragraph{Qualitative comparison.}
\Fref{fig:result_comparison} illustrates the qualitative comparison between LSMI-U \cite{kim2021large} and our method AID. For better visibility, we apply the following post-processing: 1) convert the result images in the top three rows and input images in the bottom two rows to the sRGB color space, and 2) scale down the green channel of the illumination maps in the bottom two rows.
Our model generates more natural and ground truth-like WB results and illumination maps compared to the LSMI-U.
We contribute the improvement of AID to the model design where the final illumination maps are generated under the condition of the physical image model \Eref{eq1:image_model}, and also to the proposed loss function that matches the predictions to the proper ground truths.

We also provide the plots of chromaticity of the pixel-wise illumination predictions with the ground truths. 
Through the ground-truth illumination distributions of the top three rows, illustrated in red, it can be confirmed that each scene has a single, dual, or triple illuminant, respectively represented as a point, a line segment, and a triangle.
It can be easily notified that the previous model generates unrefined predictions whereas our model performs well on reconstructing the actual distribution of the chromaticities of the illumination.
For additional visualizations, including results related to the MIIW dataset, please refer to the supplementary material.

\subsection{Generalization using single-illuminant DB}
We also assessed the generalizability of our framework using the established single-illuminant white balance dataset, NUS-8 \cite{cheng2014illuminant}, which is a well-known benchmark widely used in the literature. NUS-8 dataset comprises 8 camera subsets, and we conducted three-fold cross-validation experiment for each camera. We measured the following metrics in the same way as previous studies \cite{barron2015convolutional, barron2017fast, hu2017fc4}: mean, median, tri-mean, best 25\%, worst 25\%, and their geometric mean (G.M.). 
For the model configuration, we use $K = 5$, $T = 3$, and $D_{slot}, D_{attn} = 64$.

In \Tref{tab:generalization}, we report the angular error in degrees, along with the performance of recent works. The result shows that the proposed framework works robustly under both single- and multi-illuminant environments.

\begin{table}
\begin{center}
\resizebox{0.95\linewidth}{!}{
\begin{tabular}{lcccccc}
\toprule
  \textbf{NUS-8 \cite{cheng2014illuminant}}&
  Mean &
  Med. &
  Tri. &
  \begin{tabular}[c]{@{}l@{}}Best\\ 25\%\end{tabular} &
  \begin{tabular}[c]{@{}l@{}}Worst\\ 25\%\end{tabular} &
  G.M. \\ \midrule \midrule
CCC \cite{barron2015convolutional} &
  2.38 &
  1.48 &
  1.69 &
  0.45 &
  5.85 &
  1.73 \\
AlexNet-FC4 \cite{hu2017fc4} &
  2.12 &
  1.53 &
  1.67 &
  0.48 &
  4.78 &
  1.66 \\
FFCC \cite{barron2017fast} &
  1.99 &
  1.31 &
  1.43 &
  \textbf{0.35} &
  4.75 &
  1.44 \\
CLCC \cite{lo2021clcc} &
  1.84 &
  1.31 &
  1.42 &
  0.41 &
  4.2 &
  1.42 \\
AID &
  \textbf{1.57} &
  \textbf{1.03} &
  \textbf{1.16} &
  0.37 &
  \textbf{3.67} &
  \textbf{1.21}
  \\ \bottomrule
\end{tabular}}
\end{center}
\vspace{-1.5em}
\caption{Three-fold cross-validation result on NUS-8 dataset, with mean angular error in degrees.}
\label{tab:generalization}
\end{table}

\begin{table}
\begin{center}
{\small
\resizebox{0.7\linewidth}{!}{
\begin{tabular}{cccc}
\toprule
           & \multirow{2}{*}{\# of illum acc.} & \multicolumn{2}{c}{illuminant AE} \\ \cmidrule(lr){3-4}
           &                                   & mean            & median           \\ \midrule \midrule
Galaxy     & 0.800                               & 1.71             & 1.25              \\
Sony       & 0.871                                & 1.50              & 0.86                \\
Nikon      & 0.813                                 & 1.84               & 1.20                \\
\bottomrule
\end{tabular}
}
}
\end{center}
\vspace{-1.5em}
\caption{Additional validation metrics. We measured 1) the accuracy of predicted number of illuminants in the scene and 2) the angular error (AE) between predicted and the GT chromaticities.}
\label{tab:additional_metric}
\end{table}

\subsection{Fully decomposed multi-illuminant WB}
Since our model generates fully decomposed illumination map, we can calculate the number of illuminants or the prediction accuracy of individual illuminant's chromaticity.

\paragraph{Count \& chromaticity prediction result.}
We can also evaluate the accuracy of the predicted number of illuminants in the scene and the angular error of the chromaticities of each individual illuminant, using decomposed illumination map.
The number of illuminants was measured by ignoring slots where the maximum value of the weight map component was below the threshold of 0.3. Angular error of illuminant chromaticity was measured between chromaticity vectors with matched indices $\sigma_m$ and their corresponding GT vectors.
Such additional information could be utilized as an additional metric for how well the model understands and accurately decomposes the multi-illuminant scene. \Tref{tab:additional_metric} demonstrates that AID accurately predicts the chromaticity and the number of each illuminant. One thing to note is that it is impossible to measure this decomposition performance in previous works as ours is the first work to enable this illumination decomposition in multi-illuminant scenes.

\paragraph{Controllable multi-illuminant WB.}
Unlike previous multi-illuminant WB methods, AID can make fully-decomposed results with illuminant-wise chromaticities and weight maps.
Therefore, we can leverage these decomposed information to provide additional features like manipulating the chromaticity of each light or selective WB.
\Fref{fig:selective_awb} shows additional capabilities of AID framework.

\subsection{Ablation study}
As shown in the \Tref{tab:ablation}, we present three different ablation studies: centroid-matching loss ($\mathcal{L}_m$), the number of slots ($K$), and the number of iterations in the slot attention module ($T$).
The first section of the table shows that the centroid-based matching loss helps the model to decompose the mixed illumination with the proper number of slots, as demonstrated by the illuminant number prediction accuracy (\# acc.). 
Absence of $\mathcal{L}_{centroid}$ resulted in failure to effectively decompose mixed illumination using slots, as all slots were indiscriminately engaged to estimate the illumination, yielding an decomposition accuracy of 0.288. The efficacy of the centroid matching loss is more clearly demonstrated in Section C and Fig. 11 of the supplementary material.
In addition, the second and the third section of the study reveals that the model performance can deliver different results depending on the number of slots ($K$) and the number of iterations in the slot attention module ($T$).
\begin{table}
\begin{center}
\resizebox{\linewidth}{!}{
\begin{tabular}{cccccccc}
\toprule
\multirow{2}{*}{$\mathcal{L}_{mixed}$} & \multirow{2}{*}{$\mathcal{L}_{centroid}$} & \multirow{2}{*}{$K$} & \multirow{2}{*}{$T$} & \multicolumn{2}{c}{Mixed illum MAE} & \multicolumn{2}{c}{Illuminant} \\ \cmidrule(lr){5-6} \cmidrule(lr){7-8}
  &   &                &        & mean & median & \# acc. & AE \\ \midrule \midrule
\checkmark &   & 7 & \multicolumn{1}{c|}{3} & 1.58 & 1.26   & 0.288         & -         \\
\checkmark & \textbf{\checkmark} & 7 & \multicolumn{1}{c|}{3} & 1.66 & 1.41   & 0.800           & 1.71      \\ \hline
\checkmark & \checkmark & \textbf{5} & \multicolumn{1}{c|}{3} & 1.82 & 1.37   & 0.744         & 2.40      \\
\checkmark & \checkmark & \textbf{7} & \multicolumn{1}{c|}{3} & 1.66 & 1.41   & 0.800           & 1.71 \\
\checkmark & \checkmark & \textbf{9} & \multicolumn{1}{c|}{3} & 1.84 & 1.42   & 0.488          & 1.49     \\ \hline
\checkmark & \checkmark & 7 & \multicolumn{1}{c|}{\textbf{2}} & 1.85 & 1.46   &  0.688  & 1.79      \\
\checkmark & \checkmark & 7 & \multicolumn{1}{c|}{\textbf{3}} & 1.66 & 1.41   & 0.800           & 1.71 \\
\checkmark & \checkmark & 7 & \multicolumn{1}{c|}{\textbf{4}} & 1.92 & 1.44   &  0.720  & 1.84      \\
\bottomrule
\end{tabular}}
\end{center}
\vspace{-1.5em}
\caption{Results of ablation studies on the centroid-matching loss ($\mathcal{L}_m$), the number of slots ($K$), and the number of iterations of GRU ($T$) using the Galaxy camera subset of the LSMI dataset.}
\vspace{-1.5em}
\label{tab:ablation}
\end{table}

Among the combinations of $K$ and $T$, we choose to use $K=7$ and $T=3$ combination by considering the accuracy and the computational cost.
Ablation studies are conducted using the Galaxy camera subset of the LSMI dataset.
\section{Conclusion and discussion}

In this paper, we introduced a framework called AID, designed to extract the chromaticity of individual illuminants along with their corresponding weights, while satisfying the linearity constraint of the Lambertian image model. 
To construct our model, we incorporated the slot attention module and applied the centroid-based matching loss, extending upon previous multi-illuminant white balance methods.

We demonstrated the effectiveness of AID through various experiments, and we believe this marks as a step towards more interpretable image enhancement, particularly in the context of white balancing. 
However, we acknowledge limitations in our proposed method, such as the requirement for presets regarding the number of clusters. 
Building model that can dynamically determine the number of clusters based on input images can be a promising path for future research.

{
    \small
    \bibliographystyle{ieeenat_fullname}
    \bibliography{main}
}

\end{document}


\maketitlesupplementary
\appendix

\section{More Qualitative Results}
\begin{figure*}[t]
\begin{center}
    \includegraphics[width=0.9\textwidth]{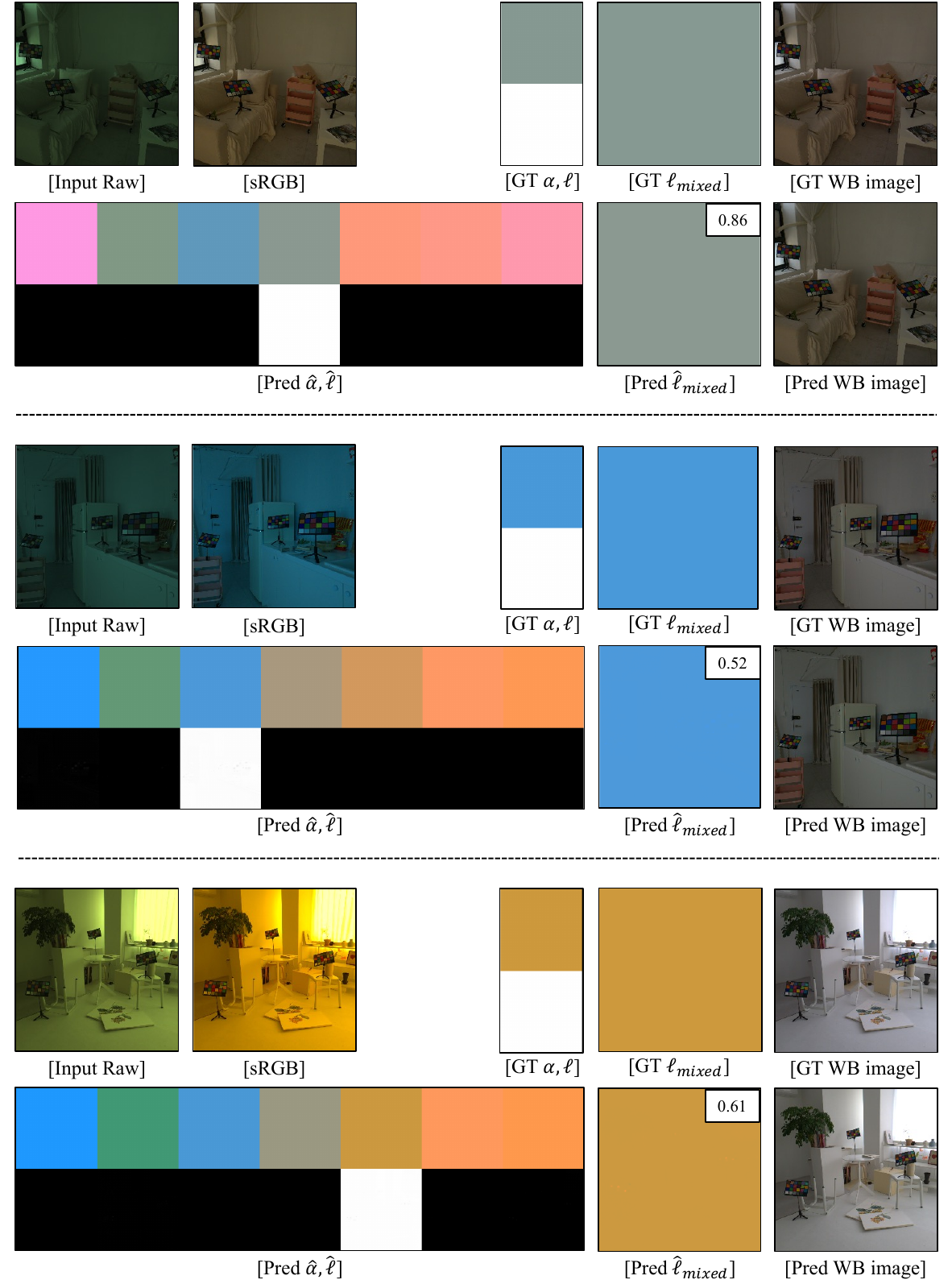}
    \caption{Single-illuminant WB result of LSMI Galaxy test set.}
\label{fig:gal_1}
\end{center}
\end{figure*}

\begin{figure*}[t]
\begin{center}
    \includegraphics[width=0.9\textwidth]{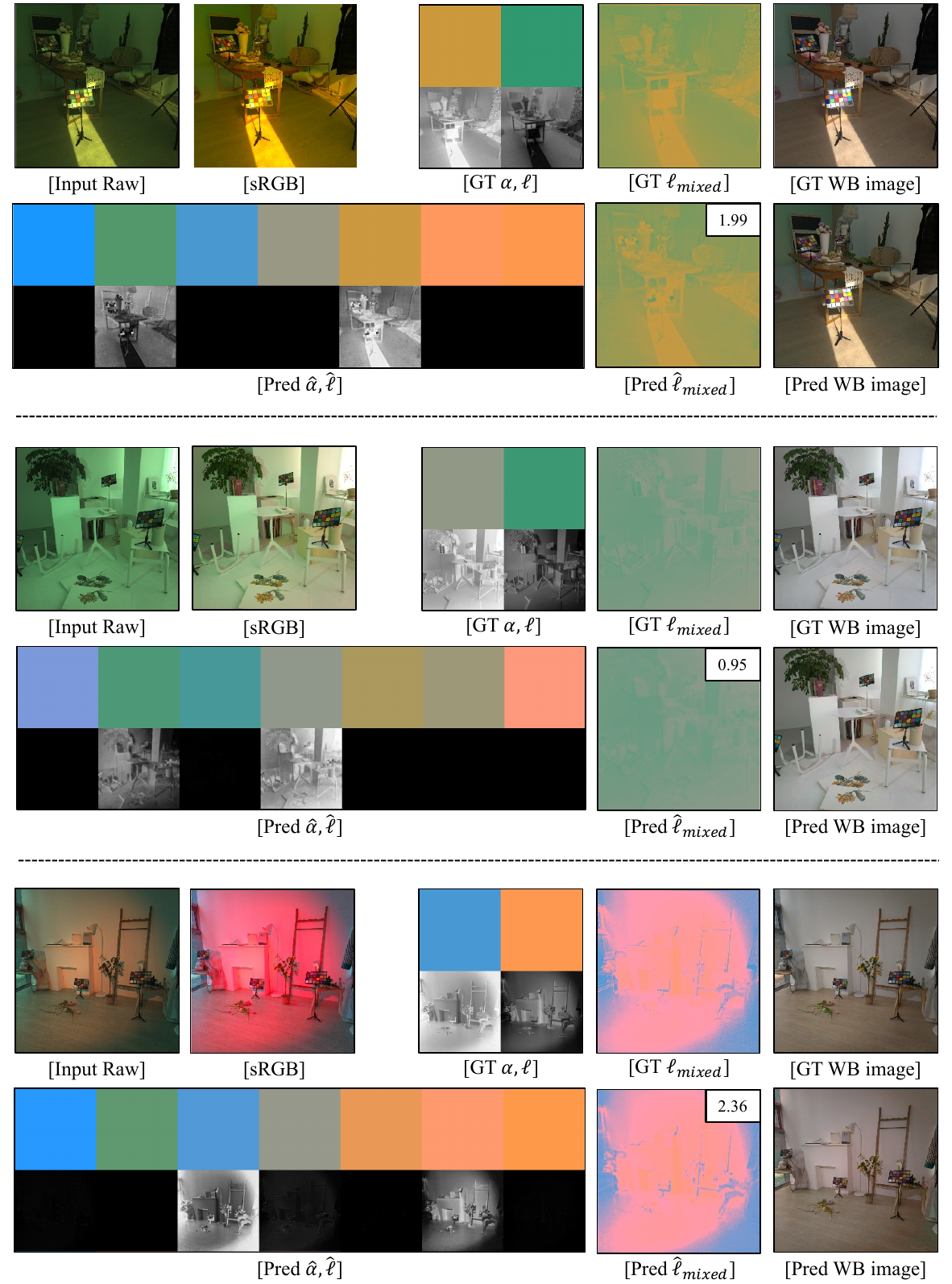}
    \caption{Multi-illuminant WB result of LSMI Galaxy test set.}
\label{fig:gal_2}
\end{center}
\end{figure*}

\begin{figure*}[t]
\begin{center}
    \includegraphics[width=0.9\textwidth]{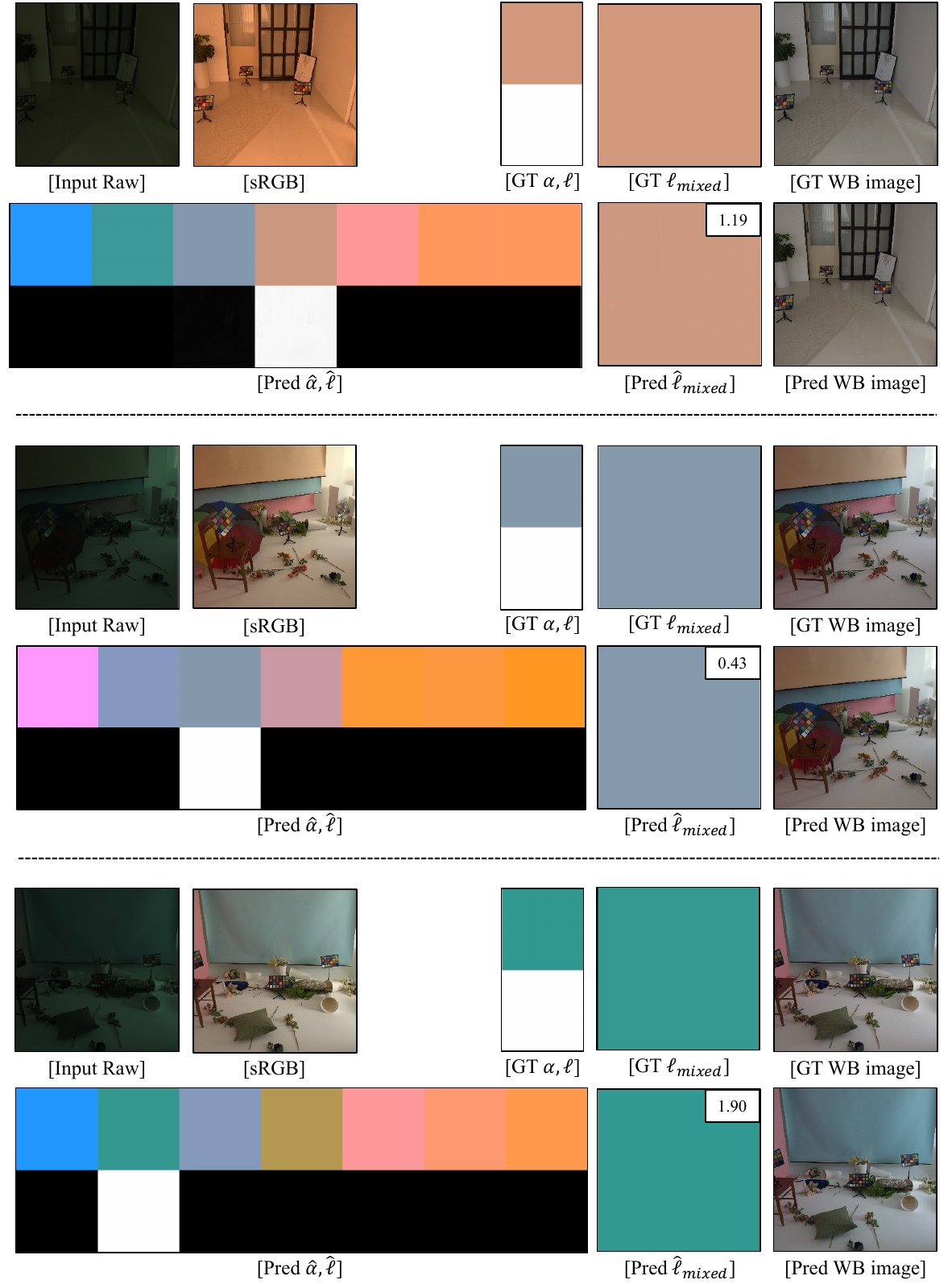}
    \caption{Single-illuminant WB result of LSMI Nikon test set.}
\label{fig:nik_1}
\end{center}
\end{figure*}

\begin{figure*}[t]
\begin{center}
    \includegraphics[width=0.9\textwidth]{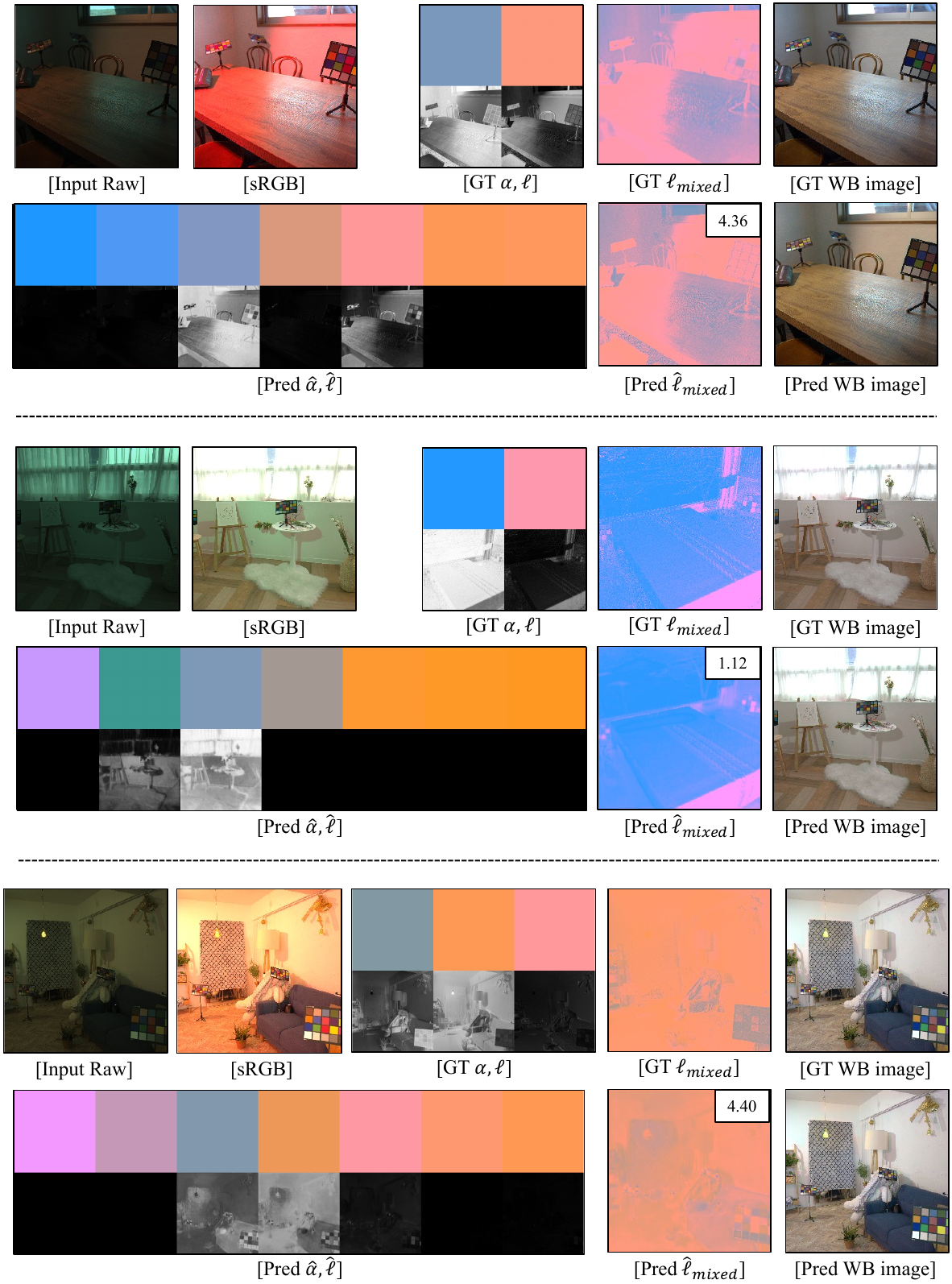}
    \caption{Multi-illuminant WB result of LSMI Nikon test set.}
\label{fig:nik_2}
\end{center}
\end{figure*}

\begin{figure*}[t]
\begin{center}
    \includegraphics[width=0.9\textwidth]{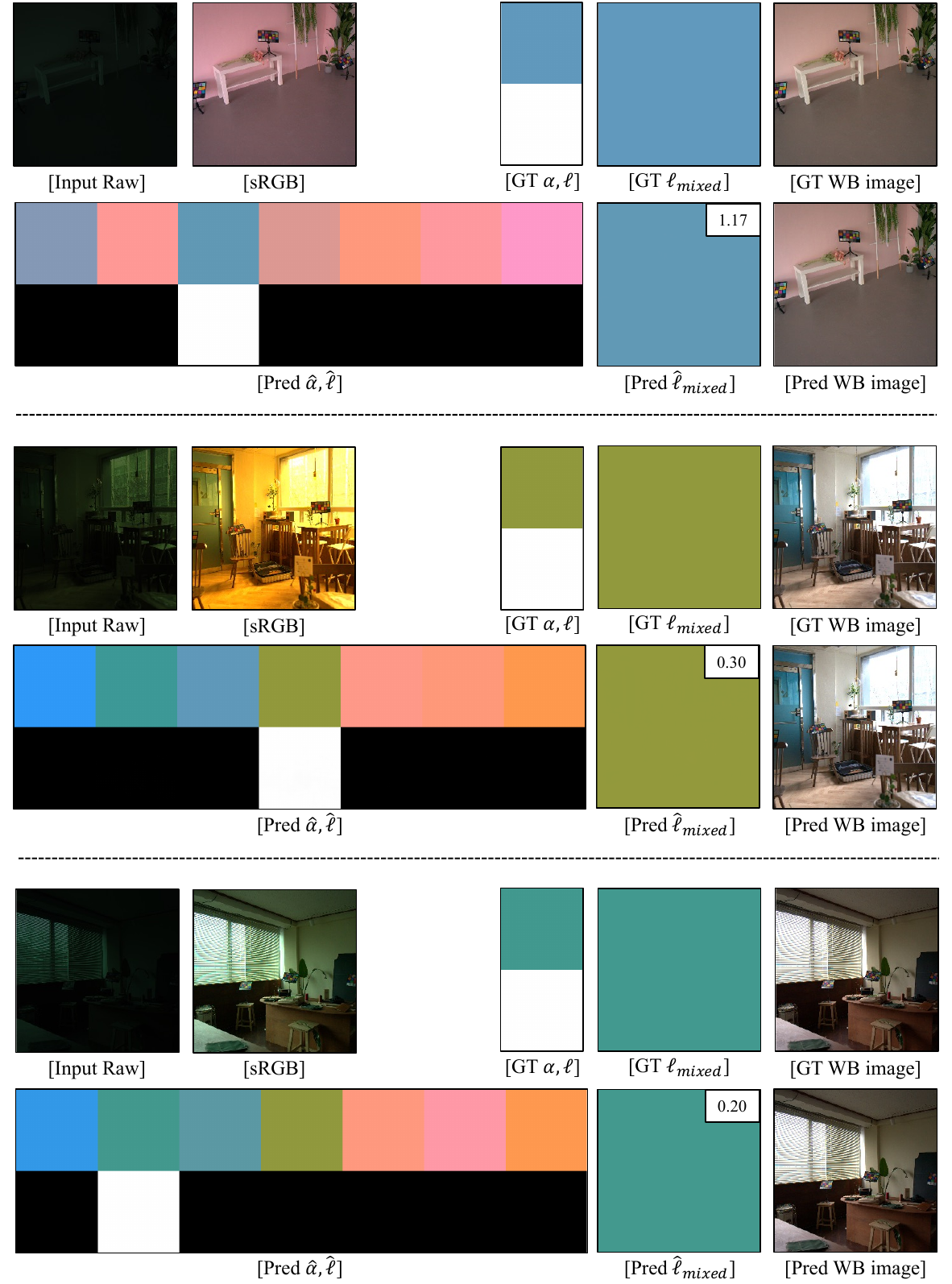}
    \caption{Single-illuminant WB result of LSMI Sony test set.}
\label{fig:son_1}
\end{center}
\end{figure*}

\begin{figure*}[t]
\begin{center}
    \includegraphics[width=0.9\textwidth]{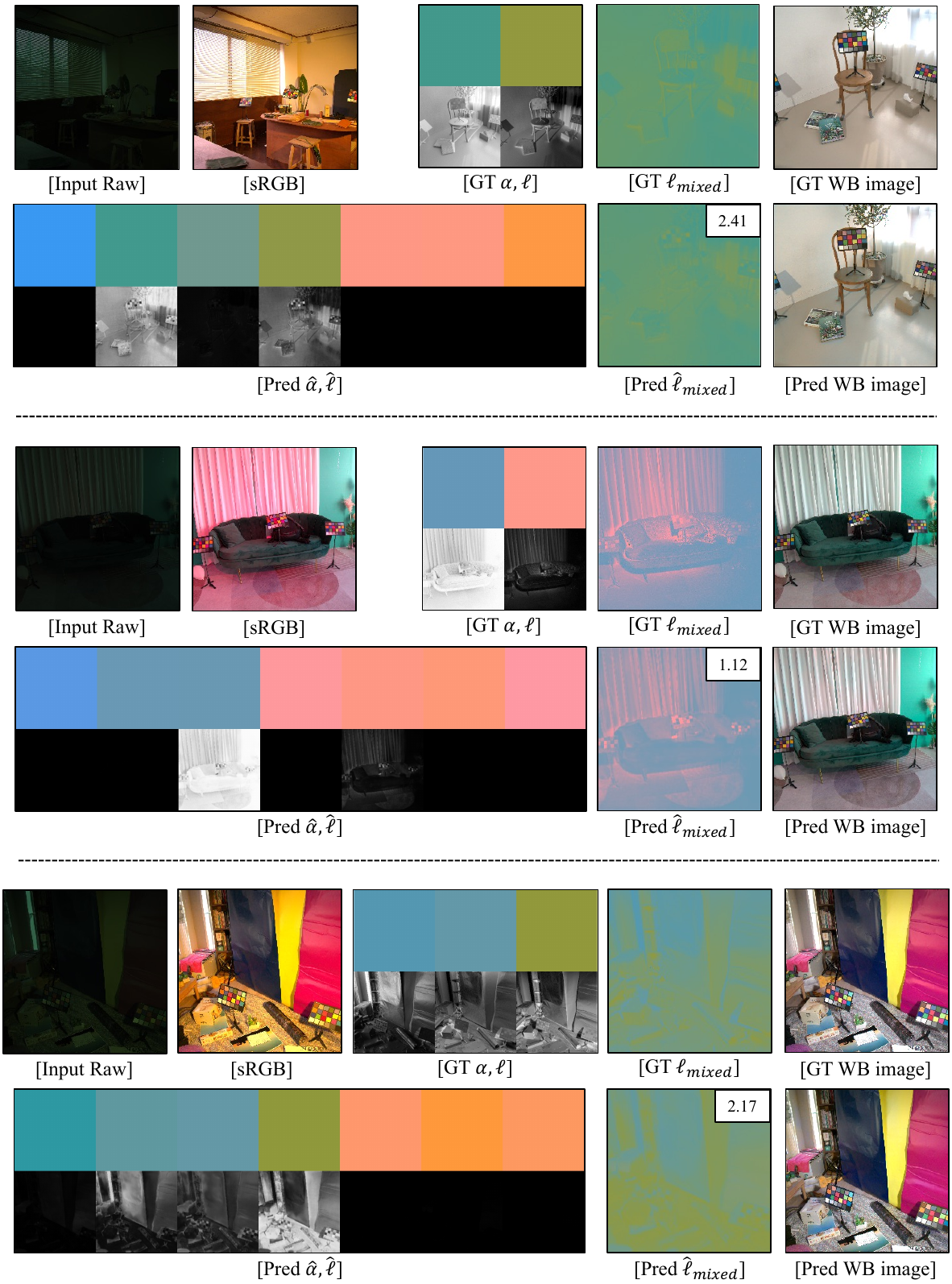}
    \caption{Multi-illuminant WB result of LSMI Sony test set.}
\label{fig:son_2}
\end{center}
\end{figure*}

\begin{figure*}[t]
\begin{center}
    \includegraphics[width=0.9\textwidth]{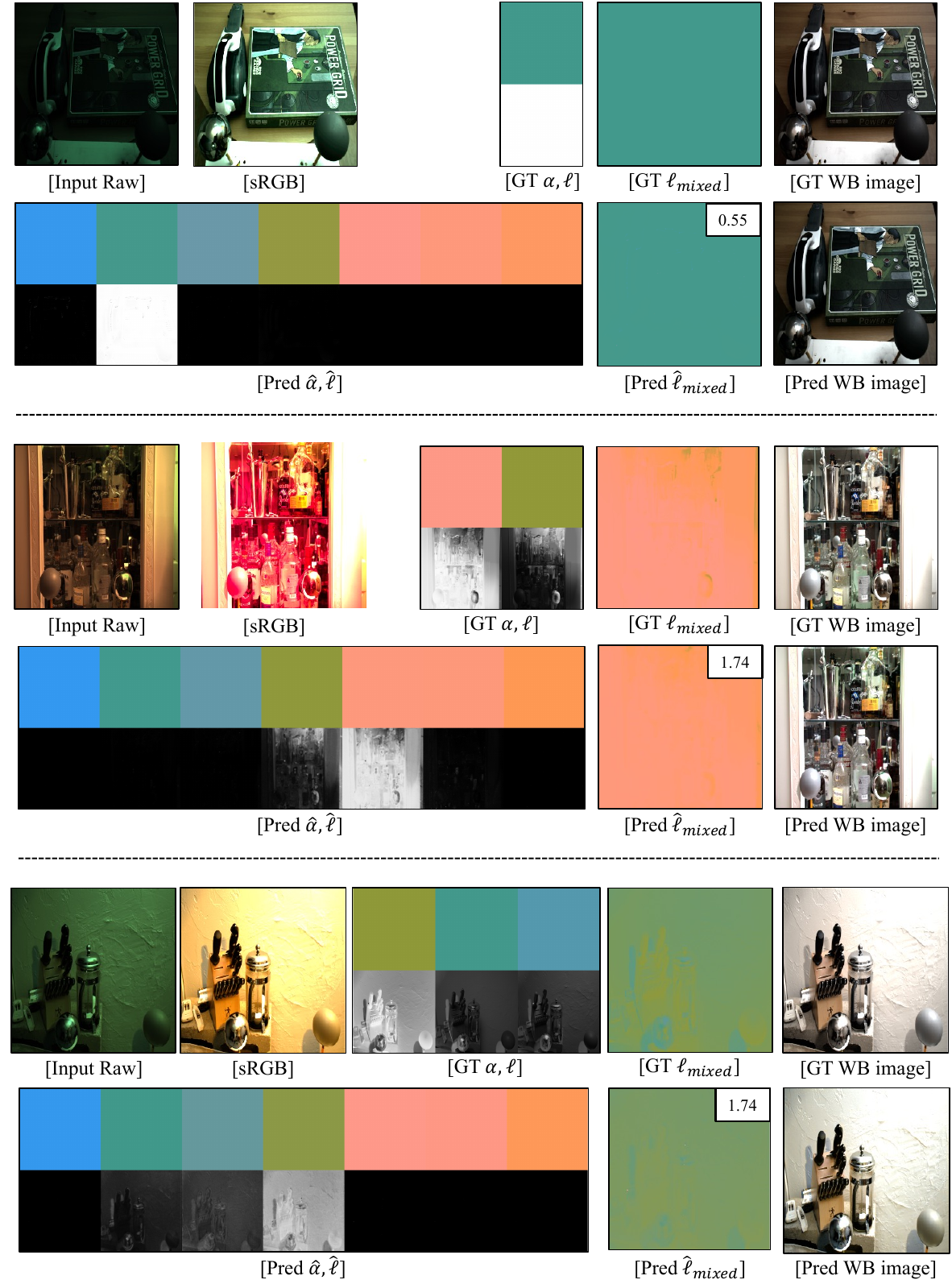}
    \caption{WB results of MIIW dataset (1,2,3 illuminant scene) using AID framework.}
\label{fig:miiw_result}
\end{center}
\end{figure*}

\begin{figure*}[t]
\begin{center}
    \includegraphics[width=\textwidth]{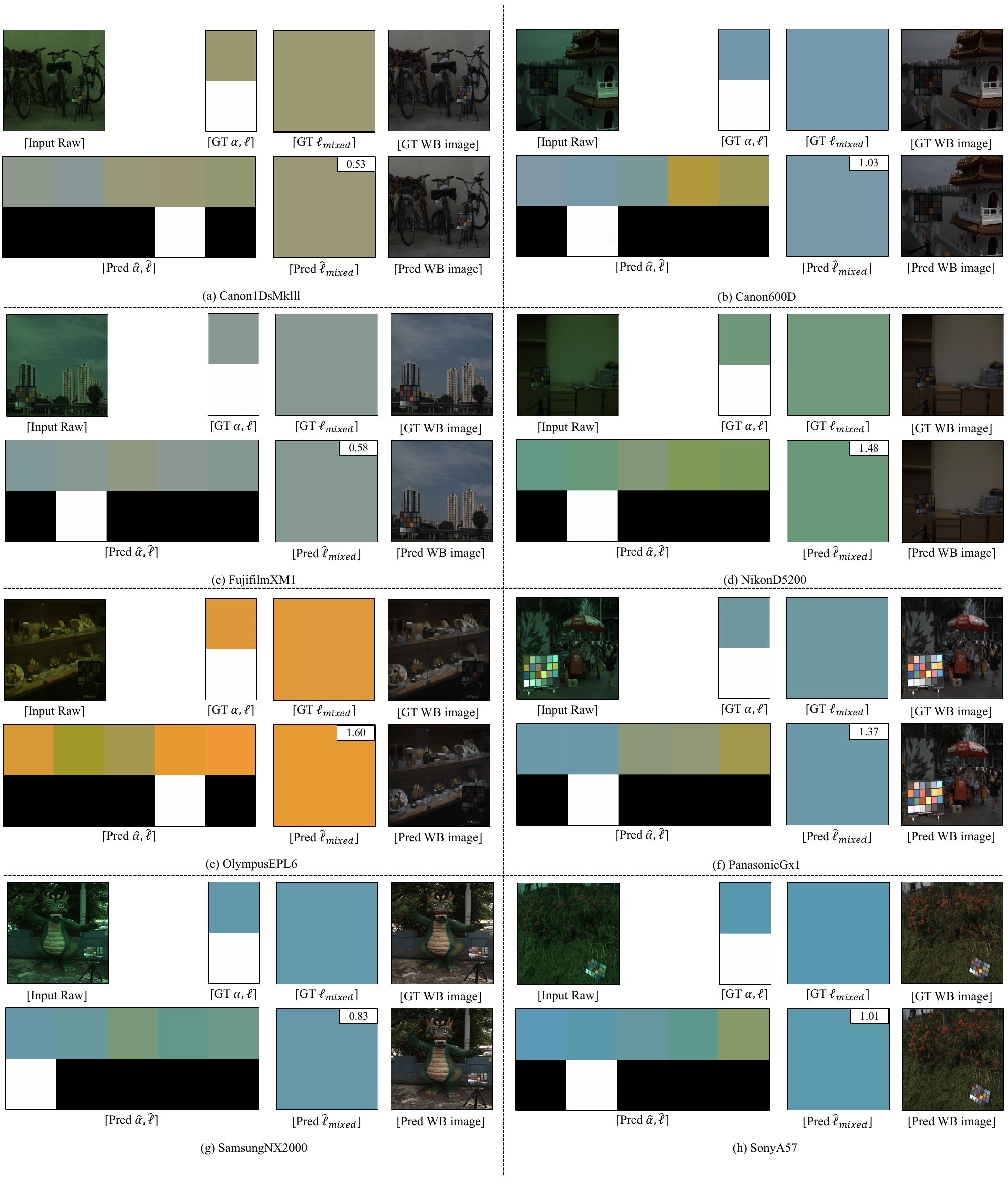}
    \caption{WB results of NUS-8 dataset using AID framework (5 slots are used).}
\label{fig:nus8}
\end{center}
\end{figure*}

More qualitative results are shown in this section. We include the prediction results of all the slots that could not be included in the paper due to spatial constraints. As noted in the paper, we used seven slots for the multi-illuminant WB experiment on the LSMI \cite{kim2021large} and MIIW \cite{murmann2019dataset} dataset and five slots for single-illuminant WB on the NUS-8 dataset \cite{cheng2014illuminant}. The LSMI galaxy subset results are presented in \Fref{fig:gal_1} and \ref{fig:gal_2}, while \Fref{fig:nik_1}, \ref{fig:nik_2}, and \Fref{fig:son_1}, \ref{fig:son_2} display the Nikon and Sony results, respectively. The angular errors between the ground truths and the predicted results are indicated in degrees at the upper right corner of the illumination maps. \Fref{fig:nus8} shows the test result of NUS-8. \textcolor{black}{ \Fref{fig:miiw_result} shows a visualization of the prediction results for MIIW dataset. Each row in the figure shows the results for a 1,2,3-illuminant scene, respectively. All visualizations demonstrate that each slot is effectively activated by the color gamut assigned to it through centroid-matching loss, while the remaining slots remain inactive. We emphasize that the AID framework demonstrates consistent and robust predictions of illuminant chromaticity and weight maps across the LSMI, MIIW, and NUS-8 datasets.}


\section{Slot Evolution through Iterative Calibraion}
\begin{figure*}[t]
\begin{center}
    \includegraphics[width=0.9\textwidth]{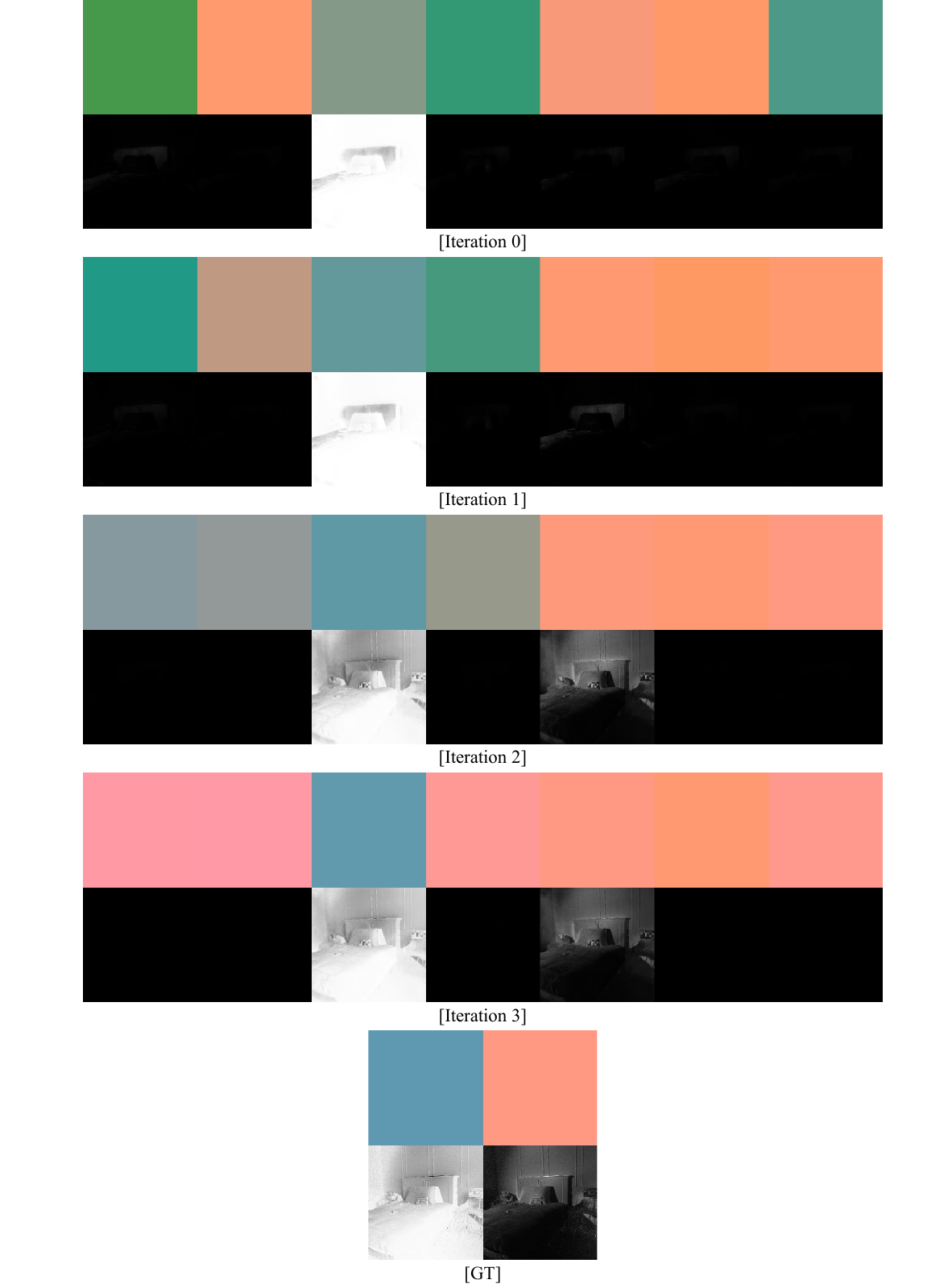}
    \caption{Slot evolution visualization with respect to iterative calibration process.}
\label{fig:evolution}
\end{center}
\end{figure*}

We also visualize the gradual evolution results of every slot and weight map. As shown in \Fref{fig:evolution}, the chromaticity and weight map of each slot become more accurate as the iteration progresses.


\section{Importance of Centroid-Matching Loss}
\begin{figure*}[t]
\begin{center}
    \includegraphics[width=0.9\textwidth]{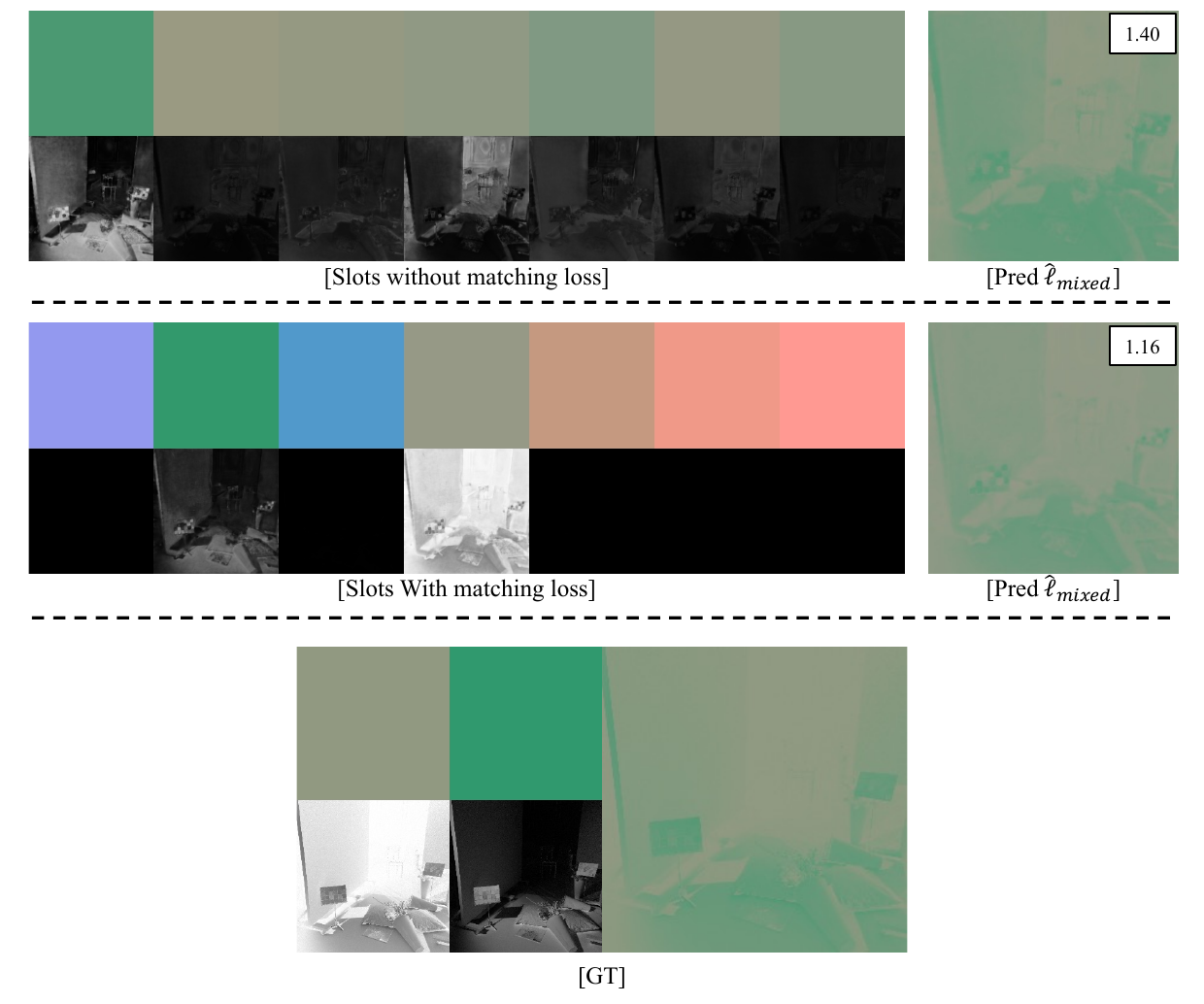}
    \caption{Without the matching loss $\mathcal{L}_{centroid}$, each slot evolves its own chromaticity to approximate the ground truth illumination, resulting in conflicts among the slots.}
\label{fig:wo_matchingloss}
\end{center}
\end{figure*}

As demonstrated in the paper, our centroid-matching loss guides the GRU to update slots more accurately. Here, we visualize the test result of AID trained without centroid-matching loss $\mathcal{L}_{centroid}$ to clarify the effect of $\mathcal{L}_{centroid}$. \Fref{fig:wo_matchingloss} shows that without the matching loss $\mathcal{L}_{centroid}$, every slot tries to engage and evolve its own chromaticity to approximate the ground truth illumination, resulting in conflicts among the slots. On the other hand, by using the matching loss, each slot specializes in a specific chromaticity range, and the slots that do not match the GT illuminant chromaticity are deactivated, resulting in more accurate illumination predictions for the scene.


\section{More Implementation Details}
For all experiments in the main paper, we use square images with a size of 256 for both training and testing.

\paragraph{AID Framework}
We only describe two components, $\mathcal{L}_{mixed}$ and $\mathcal{L}_{centroid}$, that constitute the loss term in the main paper due to the spatial constraint. The total training loss is the sum of these two terms, and it is trained through \Eref{eq:training_loss}.

\begin{equation}
    \mathcal{L}_{mixed} + \mathcal{L}_{centroid}
\label{eq:training_loss}
\end{equation}

\paragraph{Experiment on NUS-8}
To maintain consistency with previous works, we employ the same experimental setup, which involves conducting three-fold cross-validation for each camera and measuring the mean, median, tri-mean, best 25\%, and worst 25\% MAE in degrees. In line with the methodology of previous works, we report the geometric mean of each metric obtained from the eight cameras in the main paper. To augment the training data, we randomly sample ground truth illumination from a pool for each training fold and apply it to the scene during training.

\begin{figure}[t]
    \includegraphics[width=\linewidth]{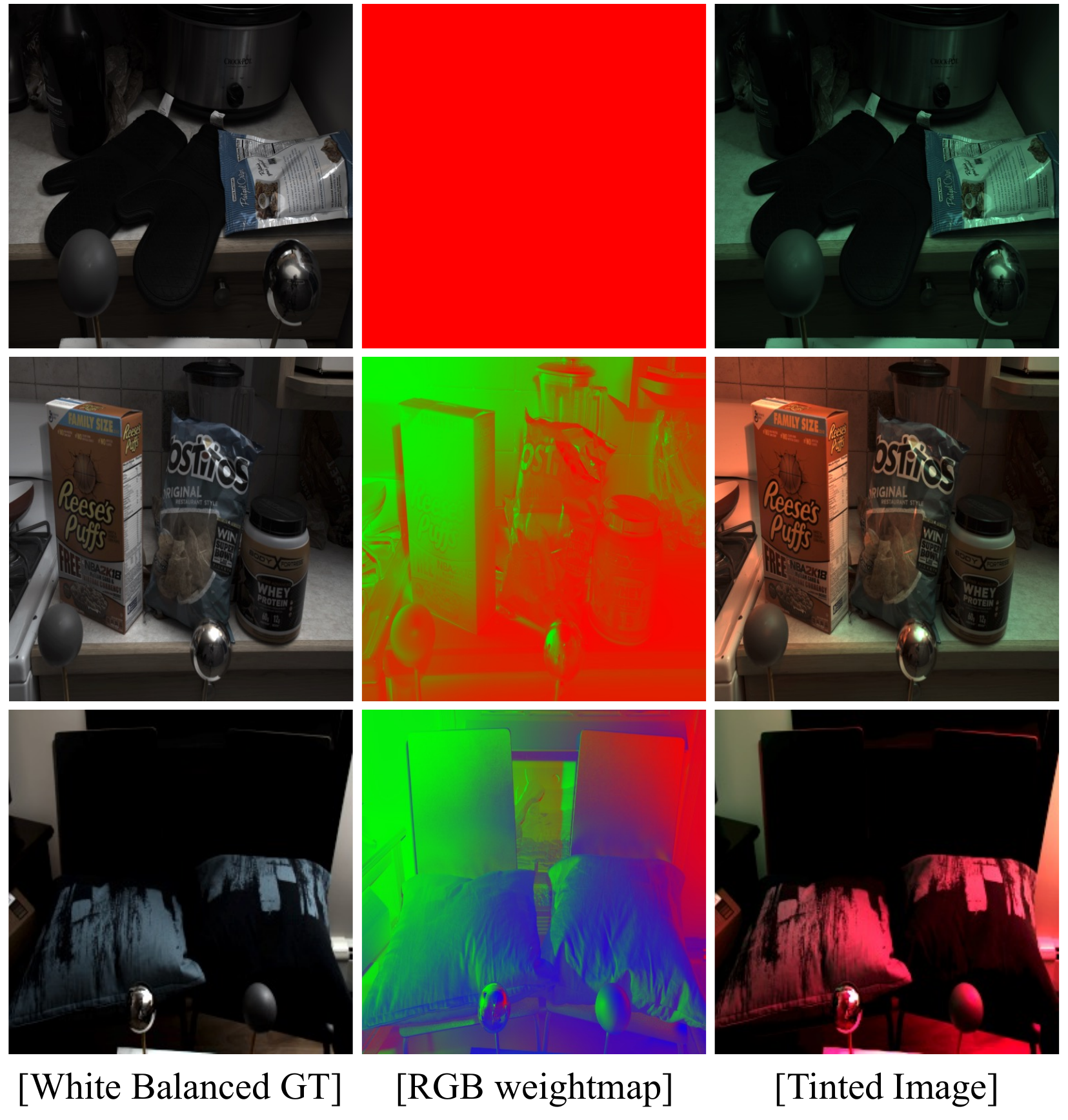}
    \caption{MIIW dataset visualization. Each row shows 1-, 2-, and 3-illuminant scene.}
    \label{fig:miiw_visualization}
\end{figure}

\textcolor{black}{\paragraph{Multi-Illumination In the Wild}
The Multi-Illumination In the Wild dataset comprises images with 25 different lighting settings for each scene, encompassing over a thousand diverse scenes. We generated train, validation, and test sets using a methodology similar to that employed in the original paper \cite{murmann2019dataset}. For each given scene, we selected one to three images to create 1- to 3-illuminant images, and the weight map $\alpha$ was generated using the G channel of the sampled images. Illuminant chromaticity was also randomly sampled from 1 to 3, corresponding to the number of sampled images, and multiplied with the weight map to create the ground truth mixed illumination map. \Fref{fig:miiw_visualization} shows generated multi-illuminant scenes, using MIIW dataset. The MIIW dataset includes canonical objects such as gray and chrome balls within the scenes. These objects were masked during training to ensure that the model did not use them as part of its learning process.}

\section{Failure Cases}
\begin{figure*}[t]
\begin{center}
    \includegraphics[width=\textwidth]{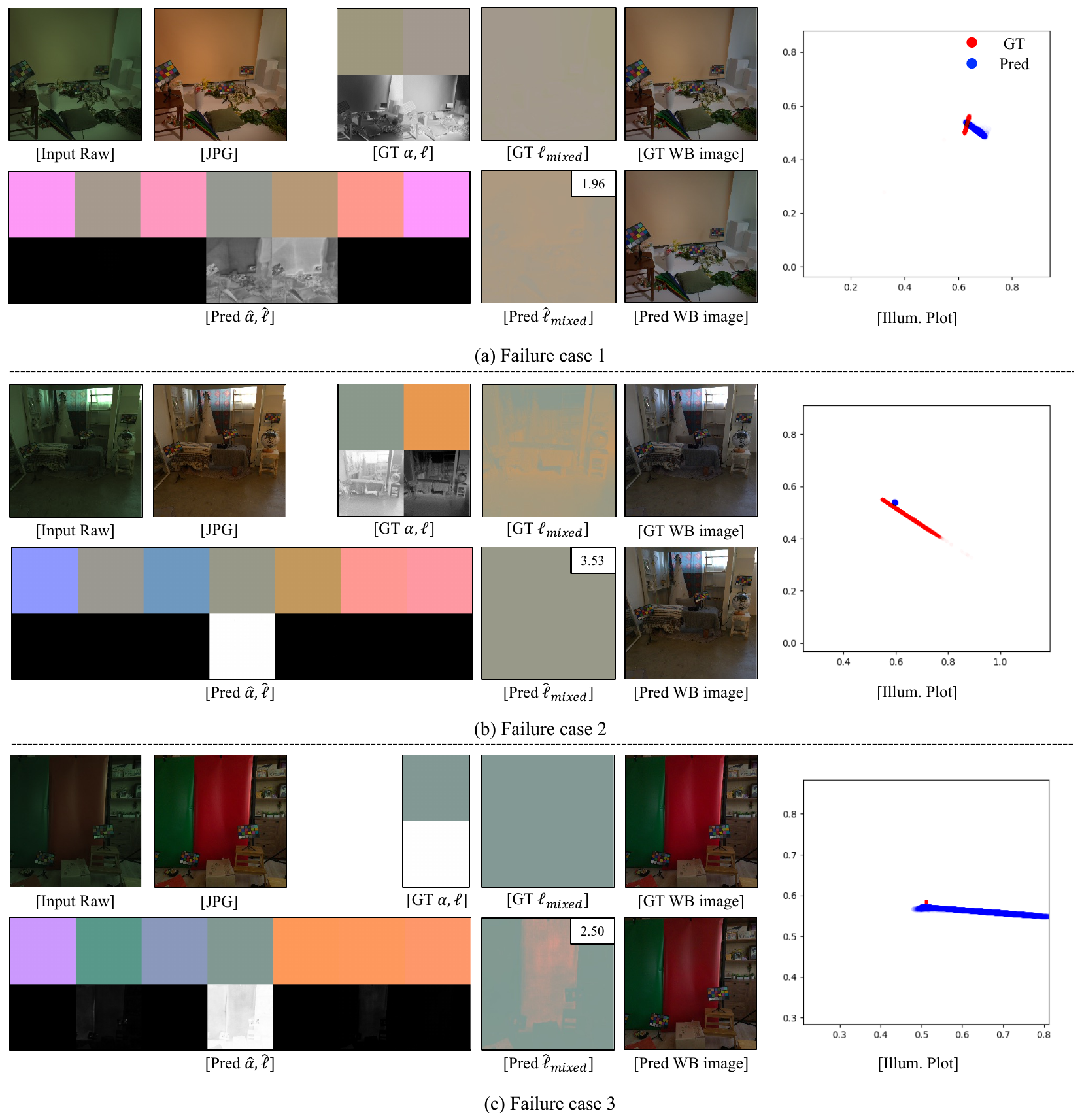}
    \caption{Failure cases of AID framework.}
\label{fig:failure}
\end{center}
\end{figure*}

Despite the strong performance of AID, we can observe some failure cases. \Fref{fig:failure}(a) illustrates that the model fails to capture each GT chromaticity for a multi-illuminant scenario. In (b), the model activates only one slot, although there are two illuminants in the scene. Lastly, (c) shows the opposite case of case 2: activates multiple slots when there is only one illuminant in a scene. We also provide the illumination plots on the rightmost of the figure to easily demonstrate the above cases.


\section{Demo Video}
We made a demo video to clearly show one of our contributions, which is an ability to controllable multi-illuminant white balance. Since AID generates a fully decomposed illumination map consisting of the chromaticity of each illuminant and a weight map, interactive WB modification is possible even in a multi-illuminant scenario. Please refer to the supplementary zip file for the video.


{
    \small
    \bibliographystyle{ieeenat_fullname}
    \bibliography{main}
}
